\documentclass[10pt,twocolumn,letterpaper]{article}

\usepackage{iccv}
\usepackage{times}
\usepackage{epsfig}
\usepackage{graphicx}
\usepackage{amsmath}
\usepackage{amssymb}

\DeclareMathOperator*{\argmin}{argmin}
\usepackage{bm}
\usepackage{algorithm}
\usepackage{algpseudocode}
\usepackage{graphics}
\usepackage{amsthm}
\usepackage{mathrsfs}
\usepackage{booktabs}
\usepackage{subfigure}
\usepackage{graphicx}
\usepackage{array}
\usepackage{multirow}
\usepackage{diagbox}
\usepackage[table,xcdraw]{xcolor}
\usepackage[font=small]{caption}

\usepackage[pagebackref=true,breaklinks=true,letterpaper=true,colorlinks,bookmarks=false]{hyperref}
\usepackage{amsfonts}

\newcommand{\ourmethod}{DenseTNT}

\newcommand{\mathbold}[1]{\mathbf{#1}}
\newcommand{\mathboldgreek}[1]{\bm{\mathrm{#1}}}
\newcommand{\heatmap}[0]{h}
\newcommand{\distance}[0]{d}

\iccvfinalcopy 

\def\iccvPaperID{7610} 

\begin{document}

\title{DenseTNT: End-to-end Trajectory Prediction from Dense Goal Sets}

\author{
Junru Gu$\phantom{}^{1}$ \hspace{15pt}
Chen Sun$\phantom{}^{2}$ \hspace{15pt}
Hang Zhao$\phantom{}^{1}$\thanks{Corresponding to: hangzhao@mail.tsinghua.edu.cn} \vspace{2pt} \\
\vspace{2pt}
$\phantom{}^1$IIIS, Tsinghua University \hspace{15pt}
$\phantom{}^2$Brown University
}

\maketitle

\begin{abstract}
Due to the stochasticity of human behaviors, predicting the future trajectories of road agents is challenging for autonomous driving.
Recently, goal-based multi-trajectory prediction methods are proved to be effective, where they first score over-sampled goal candidates and then select a final set from them.
However, these methods usually involve goal predictions based on sparse pre-defined anchors and heuristic goal selection algorithms.
In this work, we propose an anchor-free and end-to-end trajectory prediction model, named DenseTNT, that directly outputs a set of trajectories from dense goal candidates. In addition, we introduce an offline optimization-based technique to provide multi-future pseudo-labels for our final online model.
Experiments show that DenseTNT achieves state-of-the-art performance, ranking $1^{st}$ on the Argoverse motion forecasting benchmark and being the $1^{st}$ place winner of the 2021 Waymo Open Dataset Motion Prediction Challenge.\footnote{Project page: \url{https://tsinghua-mars-lab.github.io/DenseTNT}}

\end{abstract}


\section{Introduction}
For a safe and smooth autonomous driving system, an essential technology is to predict the future behaviors of road participants. For example, knowing whether other vehicles intend to cut in better helps us to make brake decisions.
However, motion prediction is a highly challenging task due to the inherent stochasticity and multimodality of human behaviors.

To model this high degree of uncertainty, some approaches predict multiple future trajectories by sampling from the distribution represented by the latent variables, \eg VAEs~\cite{cvae,yuan2019diverse} and GANs~\cite{social-gan}. Other approaches generate a set of trajectories but only perform regression on the closest one during training~\cite{social-gan,lanegcn,rasterize2019}, namely using variety loss.
However, sampling-based methods cannot output the likelihood of the predicted futures and the variety loss lacks interpretability on the outputs.

\begin{figure}[tb]
\centering
\includegraphics[width=80mm]{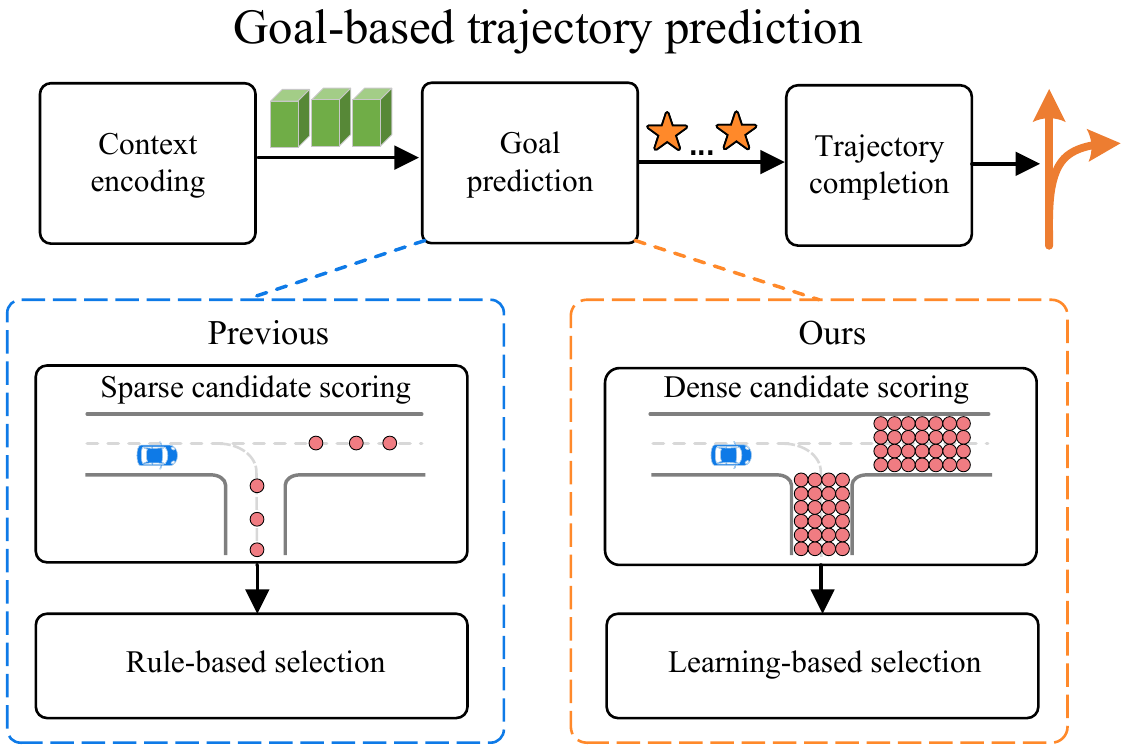}
\caption{\label{motivation}
A typical goal-based trajectory prediction pipeline is shown in the upper part of the figure.
Existing goal prediction methods~(lower left) first define sparse goal anchors heuristically, and regress and classify these anchors to estimate the goals; then rules like non-maximum suppression (NMS) are used for goal selection.  In contrast, our method~(lower right) estimates the probabilities of dense goal candidates without relying on the heuristic anchors (anchor-free). And it gets rid of rule-based post-processing by generating a set of goals in an end-to-end manner.
}
\end{figure}

More recently, goal-based methods~\cite{tnt,2021goal,lanercnn} have gained popularity and achieved state-of-the-art performance. Their key observation is that the goal (endpoint) carries most of the uncertainty of a trajectory, therefore they first predict the goal(s) of an agent, and then further complete the corresponding full trajectory for each goal.
The final goal positions are obtained by classifying and regressing predefined \textit{sparse anchors}, as shown in the lower-left part of Figure~\ref{motivation}.
For example, TNT~\cite{tnt} defines anchors as the points sampled on the lane centerlines; some others~\cite{lanercnn} take the lane segments as anchors and predicted a goal for each lane segment.
Another technique commonly adopted by these methods is to apply a rule-based algorithm to select a final small number of goals. The most notable algorithm is non-maximum suppression (NMS)~\cite{tnt}, where only locally high-scored goals are selected.

The limitations of these methods are two-folds. First, the prediction performance of these methods heavily depends on the quality of the goal anchors. Since an anchor can only generate one goal, a model cannot make multiple trajectory predictions around one anchor. Besides, sparse anchor-based methods cannot capture fine-grained information, \ie different positions on the same lane segment contain different local information, such as the relative distance to the nearest lane boundary.
Moreover, after estimating the probability of the sparse goals, NMS is used to heuristically select the goal set, which is a greedy algorithm and is not guaranteed to find the optimal solution given the multimodal nature of the problem.

\begin{figure*}[tb]
\includegraphics[width=175mm]{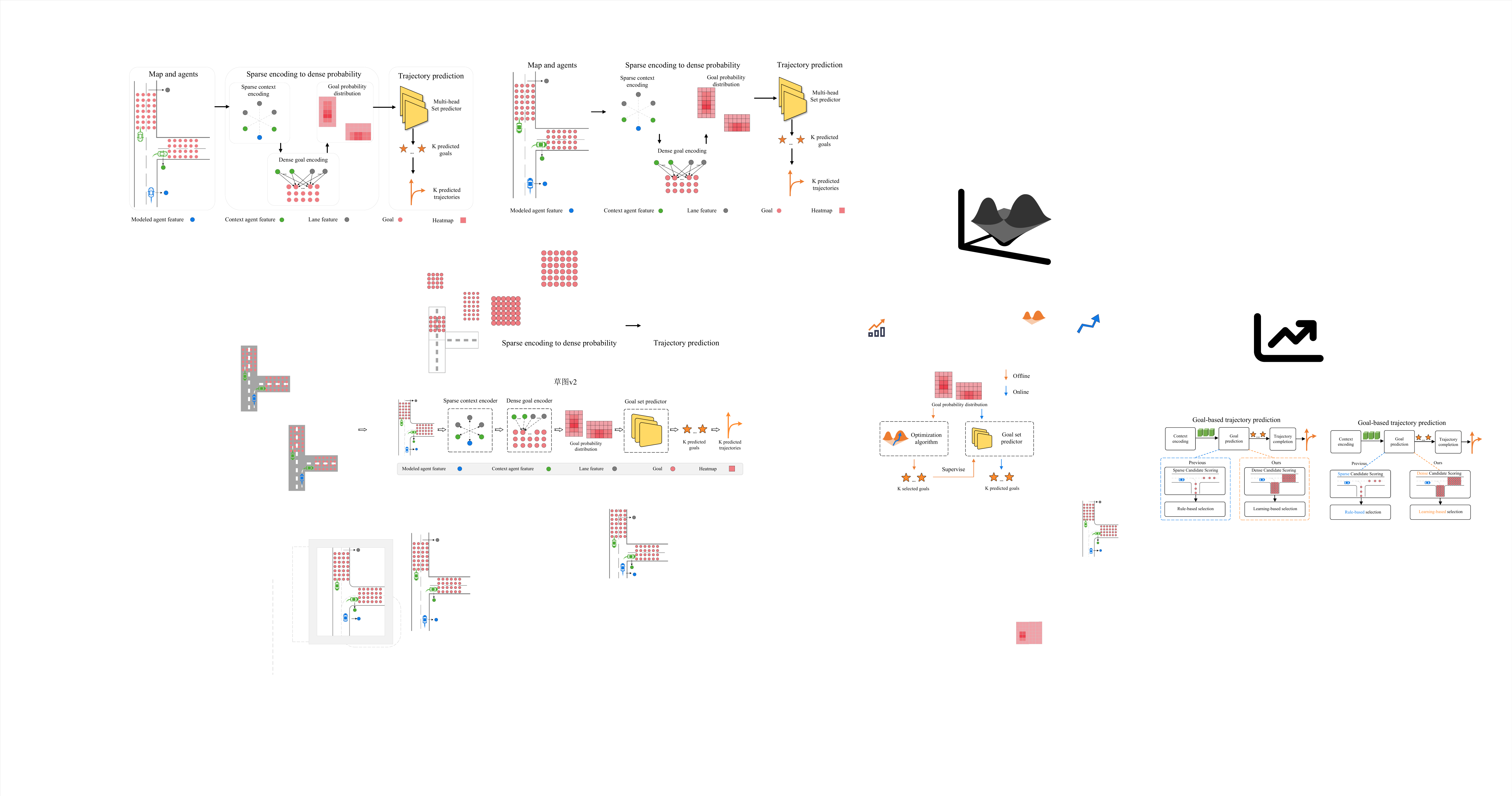}
\caption{\label{overview}
An overview of \ourmethod. A sparse context encoder is used to extract the features of HD maps and agents; and then a dense goal encoder is employed to output dense goal probability distribution; finally a goal set predictor takes the probability distribution of the goals as input and generates a set of predicted goals.
}
\end{figure*}

To address these issues, we propose DenseTNT, an anchor-free and end-to-end multi-trajectory prediction method.
DenseTNT first generates \textit{dense} goal candidates with their probabilities from the scene context; from the goal probabilities, it further employs a \textit{goal set predictor} to produce a final set of trajectory goals.
Compared to the previous methods, DenseTNT better models goal candidates and gets rid of post-processing.

Goal set prediction in DenseTNT is a multi-label prediction problem and requires multiple labels as training targets. However, unlike object detection, which innately has multiple label boxes as supervision~\cite{detr}, in trajectory prediction we only observe \textit{one} ground truth future out of many possible futures in each training sample, making it extremely challenging to supervise the model.
To tackle this problem, we devise an \textit{offline model} to provide multi-future pseudo-labels for our \textit{online model}. Compared with the above online model, the offline model uses an optimization algorithm instead of the goal set predictor for goal set prediction. The optimization algorithm finds an optimal goal set from the probability distribution of the goals; and then the set of goals are used as pseudo-labels for the training of the online model.

DenseTNT achieves state-of-the-art performance in autonomous driving trajectory prediction tasks, ranking $1^{st}$ on the Argoverse motion forecasting benchmark and $1^{st}$ in the 2021 Waymo Open Dataset Motion Prediction Challenge.

\section{Related Work}
Future predictions are highly uncertain because of the unknown intents and behaviors of agents~\cite{hasan2018mx,precog,yao2019egocentric,wimp,marchetti2020mantra,sun2020recursive,zhang2020stinet} and the complicated interactions among agents~\cite{gilles2021thomas,ngiam2021scene}. In the field of autonomous driving, to model the high degree of multimodality, implicitly using latent variables is a popular approach~\cite{hong2019rules,yeh2019diverse,sun2019stochastic,tang2019multiple}. DESIRE~\cite{cvae} used conditional variational autoencoders (CVAEs), and some approaches aimed to address mode collapse~\cite{rhinehart2018r2p2,yuan2019diverse,phan2020covernet,fang2020tpnet,Casas2020ImplicitLV}.
More recently, goal-based multi-trajectory prediction methods have gained popularity due to their superior performance. We will discuss their details later in this section.

\paragraph{Map encoding.}
The map encoding methods can be divided into two categories: rasterized encoding and vectorized encoding. Rasterized encoding methods rasterize the HD map elements together with agents into an image and use CNNs to encode the image. Based on rasterized encoding, Cui~\etal~\cite{rasterize2019} went beyond a single trajectory and predicted multiple trajectories as well as their probabilities. IntentNet~\cite{intentnet} developed a detector composed of CNNs to extract features from not only raster images but also LiDAR points. Multipath~\cite{multipath} used CNNs to extracts features from raster images, then predicted the probabilities over $K$ predefined anchor trajectories and regressed offsets from the anchor states. Liang~\etal~\cite{garden} designed multi-scale location encodings and convolutional RNNs over graphs for map encoding. To capture the uncertainty in long-range human trajectory prediction, Jain~\etal~\cite{Jain2019DiscreteRF} predicted and updated discretized distribution over spatial locations. These rasterized methods cannot capture the structural information of high-definition maps and do not allow non-grid sampling of goal points due to the shape of convolutions.

Recently, sparse (vectorized) encoding methods, which can better capture the structural information of high-definition maps, have developed rapidly. They treat each entity (a lane or an agent) as a sparse set of elements and use graph neural networks to extract both the features of the entities and the interactions among different entities. VectorNet~\cite{vectornet} is the first to directly incorporate vectorized information of both the lanes and the agents. LaneGCN~\cite{lanegcn} constructed a lane graph and used graph convolutions with adjacency matrices to capture the complex topology of the lane graph. Instead of representing each agent by a feature vector, LaneRCNN~\cite{lanercnn} proposed a graph-based representation for each agent and captured the interactions among agents by modeling graph-to-graph interactions. Besides, TPCN~\cite{ye2021tpcn} employed point cloud learning strategies to model. Unlike these vectorized methods which only consider the lane centerlines or lane boundaries of the HD maps, we model dense spatial locations on the roads.

\paragraph{Goal-based trajectory prediction.}
Rehder~\etal~\cite{rehder2015goal} introduced the pedestrian’s goal as a latent variable and thus converted the prediction problem into a planning problem.
TNT~\cite{tnt} first sampled anchors from the road maps and generated trajectories conditioned on these anchors. Trajectories were then scored and non-maximum suppression (NMS) was used to select a final set of trajectories.
Similar to TNT, the decoding pipeline of LaneRCNN~\cite{lanercnn} treated a lane segment as an anchor and output each anchor's probability, then used NMS to remove duplicate goals if two predictions are too close.
DROGON~\cite{drogon} focused on a different task that the intentional destinations of individual agents are given. They created a trajectory prediction dataset to investigate the goal-oriented behavior and used a conditional VAE framework to forecast multiple possible trajectories.
The goal-based idea has also been used in finding the optimal planning policy for autonomous driving~\cite{goal-based_planning} and in human trajectory prediction~\cite{2021goal}.
Compared to the previous works, DenseTNT is an anchor-free goal-based model that can be learned in an end-to-end manner. A concurrent work HOME~\cite{gilles2021home} that used CNNs to generate a heatmap and designed greedy algorithms for goal sampling, is very similar to our dense probability estimation.

\section{Method}
\ourmethod~is an anchor-free and end-to-end trajectory prediction method that directly outputs a set of trajectories from dense goal candidates.
We first utilize a sparse (vectorized) encoding method to extract features, which captures the structural features of high-definition maps (Section \ref{sec:sparse_encoding}). Then we employ a dense goal encoder to generate the probability distribution of the goals (Section \ref{sec:dense_probability}). Finally, a goal set predictor takes the probability distribution of the goals as input and generates a set of goals directly (Section \ref{sec:set_prediction}). To train our model, more specifically the goal set predictor, we devise an optimization-based offline model which produces pseudo-labels for supervision. 

\subsection{Sparse context encoding}
\label{sec:sparse_encoding}

Scene context modeling is the first step in behavior prediction. It extracts the features of the lanes and the agents and captures the interactions among them.
Sparse encoding methods~\cite{vectornet,lanegcn} (also called vectorized methods) were proposed recently. Compared to rasterized encoding methods which rasterize the lanes and the agents into images and use CNNs to extract features, sparse encoding methods abstract all the geographic entities (\eg lanes, traffic lights) and vehicles as polylines, and better capture the structural features of high-definition maps.

We adopt VectorNet~\cite{vectornet} in this work for its outstanding performance. VectorNet is a hierarchical graph neural network composed of a subgraph module and a global graph module. The subgraph module is used to encode the features of the lanes and the agents, and the global graph module uses the attention mechanism to capture the interactions among the lanes and the agents. After context encoding, we obtain a 2D feature matrix $\mathbf{L}$, where each row $\mathbf{L}_i$ indicates the feature of the $i^{th}$ map element (\ie, a lane or an agent).

\subsection{Dense goal probability estimation}
\label{sec:dense_probability}
After the sparse context encoding, we perform probability estimation for goals on the map. TNT~\cite{tnt} defined discretized sparse anchors on the roads and then assigned probability values upon them.
Our key observation is that sparse anchors are not a perfect approximation of real probability distributions on the roads, because (1) one anchor can only generate one goal, we cannot make multiple trajectory predictions around one anchor; (2) sparse anchor-based methods cannot capture fine-grained information, \ie different positions on the same lane segment contain different local information, such as the relative distance to the nearest lane boundary.

Therefore, we perform dense goal probability estimation on the map instead, so that the goal prediction is anchor-free. Concretely, a  dense goal encoder is used to extract the features of the locations on the road under a certain sampling rate. Then, the probability distribution of the dense goal candidates is predicted. 

\paragraph{Lane scoring.}
Before goal probability estimation, we adopt a lane scoring module to predict the lane the goal will land on to reduce the number of goal candidates. As a higher level of abstraction, there are tens of goals on each lane. By scoring lanes, we can filter away the goal candidates which are not located on the candidate lanes, reducing computation in the later stage.

The scoring of lanes is modeled as a classification problem, and a binary cross-entropy loss $\mathcal{L}_{\rm lane}$ is used for training.
The ground truth score of the lane closest to the ground truth goal is $1$, and the others are $0$. The distance between a lane $\mathbf{l}$ and the ground truth goal $y_{\rm gt}$ is defined as $\distance(\mathbf{l}, y_{\rm gt}) = \min(||l_1-y_{\rm gt}||^2,||l_2-y_{\rm gt}||^2,\dots,||l_t-y_{\rm gt}||^2)$.

\begin{figure}[t]
\centering
\includegraphics[width=80mm]{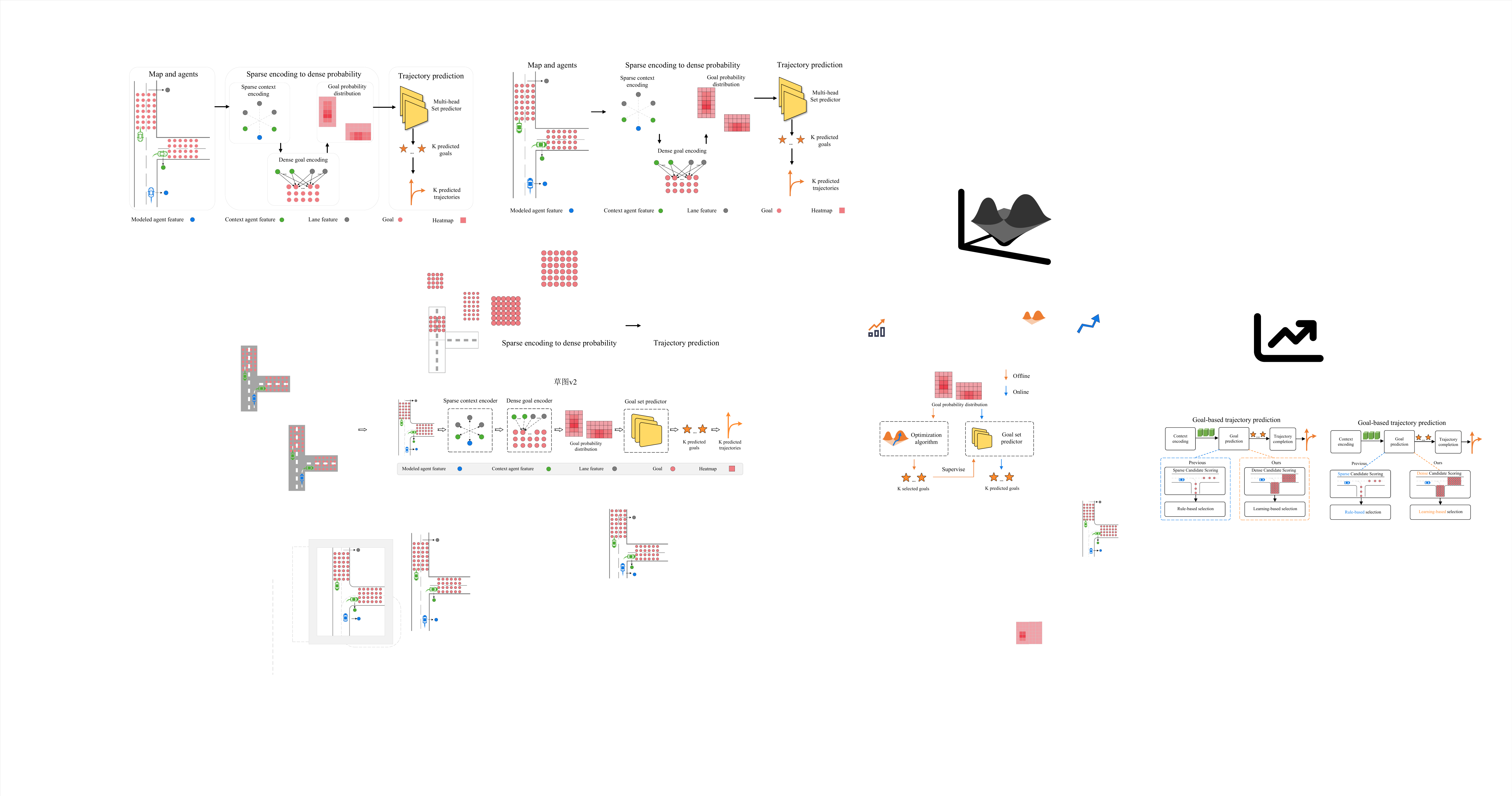}
\caption{\label{train}
Two-stage training of goal set predictor. In the first stage, we use the ground truth goals to train all the modules except for the goal set predictor. In the second stage, we only train the goal set predictor, using pseudo-labels generated by the optimization algorithm.
}
\end{figure}

\paragraph{Probability estimation.}
The  dense goal encoder uses an attention mechanism to extract the local information between the goals and the lanes.
We first get the initial feature matrix $\mathbf{F}$ of the goals by encoding their 2D coordinates using MLP.
The local information between the goals and the lanes can be obtained by attention mechanism:
\begin{equation}
    \mathbf{Q} = \mathbf{F}\mathbf{W}^Q, \mathbf{K} = \mathbf{L}\mathbf{W}^K, \mathbf{V} = \mathbf{L}\mathbf{W}^V,
\end{equation}
\begin{equation}
A \left( \mathbf{Q},\mathbf{K},\mathbf{V} \right) ={\rm softmax} \left( \frac{\mathbf{Q}\mathbf{K}^\top}{\sqrt{d_k}}\right)\mathbf{V},
\end{equation}
where $\mathbf{W}^Q, \mathbf{W}^K, \mathbf{W}^V \in \mathbb{R}^{d_h\times d_k} $ are the matrices for linear projection, $d_k$ is the dimension of query / key / value vectors, and $\mathbf{F}, \mathbf{L}$ are feature matrices of the dense goal candidates and all map elements (\ie, lanes or agents), respectively.

The predicted score of the $i^{th}$ goal can be written as:
\begin{equation}
\phi_i = \frac{\exp(g(\mathbf{F}_i))}{\sum ^{N}_{n=1} \exp(g(\mathbf{F}_n))},
\end{equation}
where the trainable function $g(\cdot)$ is also implemented with a 2-layer MLP.
The loss term for training the sparse context encoder and the dense probability estimation is a binary cross-entropy loss between the predicted goal scores $\mathboldgreek{\phi}$ and the ground truth goal scores $\mathboldgreek{\psi}$:
\begin{equation}
\mathcal{L}_{\rm goal} = \mathcal{L}_{\rm CE}(\mathboldgreek{\phi},\mathboldgreek{\psi}).
\end{equation}
The ground truth score of the goal closest to the final position is $1$, and the others are $0$.

\subsection{Goal set prediction}
\label{sec:set_prediction}
\begin{figure}[t]
\centering
\subfigure[Larger threshold is better.]{
\includegraphics[width=0.77\linewidth]{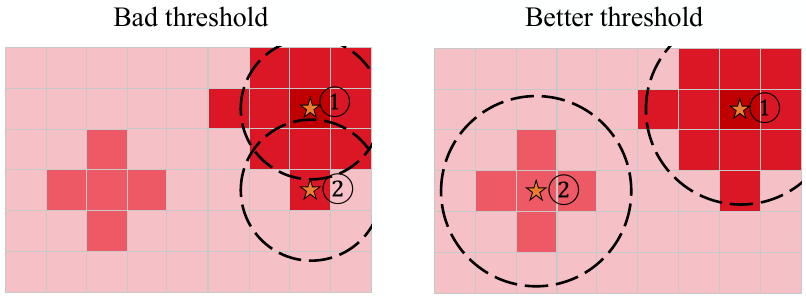}
}
\subfigure[Smaller threshold is better.]{
\includegraphics[width=0.77\linewidth]{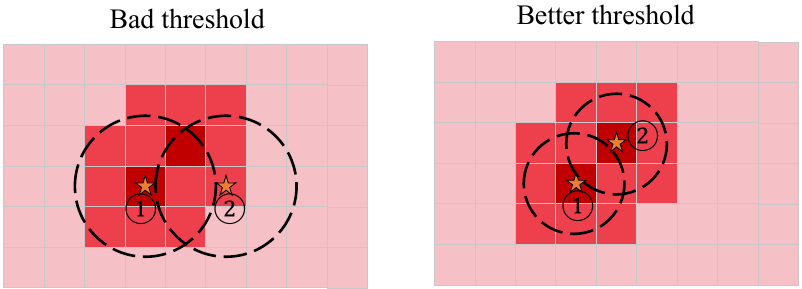}
}
\caption{NMS results in suboptimal goal selection. The upper example requires a larger threshold while the lower one needs a smaller threshold. Orange stars denote selected goals ($K=2$) of different NMS thresholds for different heatmaps, NMS threshold is depicted as the radius of the circle.}
\label{fig:nms}
\vspace{-0.15in}
\end{figure}

With the dense probability estimation above, we obtain a heatmap indicating the probability distribution of the final positions of the trajectories. We aim to pick the most likely goals across different modalities, \ie some distinctive peaks in the heatmap. Typical goal-based trajectory prediction pipeline adopts non-maximum suppression (NMS) for goal selection. However, NMS cannot handle various situations flexibly because different heatmaps have different optimal NMS thresholds, as shown in Figure~\ref{fig:nms}.

Our finding is that goal selection can be modeled as a set prediction task, so we design a goal set predictor that takes this heatmap as input and generates the goal set in an end-to-end manner.
However, different from object detection, which has multiple label boxes~\cite{detr}, in the trajectory prediction problem, we can only observe one ground truth future out of many possible futures.
To tackle this problem, we devise an offline model to provide multi-future pseudo-labels for our online model (more specifically, the goal set predictor). The offline model is composed of the same encoding modules as the online model, but with an optimization algorithm in place of the goal set predictor.
In the following, we first introduce the optimization algorithm, and then
detail our goal set predictor. The training procedure of the goal set predictor is shown in Figure~\ref{train}.

\paragraph{Optimization (offline).}
The heatmap obtained from the above steps is denoted by a mapping $\heatmap$ from $\mathcal{C}=\{c_1,c_2,\dots,c_m\}$ to $[0, 1] \subset \mathbb{R}$, where $c_i \in \mathbb{R}^2$ is the $i^{th}$ goal on the map. Let $Y$ be random variable of the coordinates of the final position, and its probability distribution satisfies $\mathbb{P}(Y = c_i) = \heatmap(c_i)$. Given a predicted goal set $\mathbold{\hat{y}}=\{\hat{y}_1,\hat{y}_2,\dots,\hat{y}_K\}$ and the ground truth goal $y_{\rm gt}$, the error of $\mathbold{\hat{y}}$ is $\distance(\mathbold{\hat{y}}, y_{\rm gt})$, \eg, the minimal final displacement error (FDE) is:
\begin{equation}
\distance_{\text{FDE}}(\mathbold{\hat{y}}, y_{\rm gt}) = \min_{y_i \in \mathbold{\hat{y}}} ||y_i - y_{\rm gt}||.
\end{equation}

Since we do not know the exact value of $y_{\rm gt}$, it is difficult to obtain the error $\distance(\mathbold{\hat{y}}, y_{\rm gt})$. However, we can use the probability distribution of $Y$ to obtain the expected error of $\mathbold{\hat{y}}$:
\begin{equation}
\mathbb{E} [\distance(\mathbold{\hat{y}}, Y)]=\sum ^{m}_{i=1} \heatmap(c_i) \distance(\mathbold{\hat{y}}, c_i).
\end{equation}

\begin{algorithm}[t]
\caption{Optimization for Offline Goal Set Prediction}
\label{algorithm:1}
\begin{algorithmic}[1]
\State \textbf{Input:} a heatmap $\heatmap$ that is a mapping from the dense goal candidates $\mathcal{C}=\{c_1,c_2,\dots,c_m\}$ to $[0, 1] \subset \mathbb{R}$, indicating the probability distribution of the final position.
\State \textbf{Objective:} to find a goal set that minimizes $f(\mathbold{y}) = \mathbb{E} [\distance(\mathbold{y}, Y)]=\sum ^{m}_{i=1} \heatmap(c_i) \distance(\mathbold{y}, c_i)$.
\State The current goal set $\mathbold{\hat{y}}=\{\hat{y}_1,\hat{y}_2,\dots,\hat{y}_K\}$ is randomly sampled from the dense goal candidates $\mathcal{C}$, and the current expected error is $f(\mathbold{\hat{y}})$.
\For{$step \in \{1..\infty\}$ }
    \If{time exceeded}
        \State break
    \EndIf
    \For{each goal $\hat{y}_i$}
        \State $\hat{y}_i^{'} = {\rm RandomPerturbation}(\hat{y}_i)$
    \EndFor
    \State $e = f(\mathbold{\hat{y}})$
    \State $e^{'} = f(\mathbold{\hat{y}}^{'})$
    \State $r={\rm Random}(0,1)$
    \If{$e < e^{'}$ or $r<0.01$}
        \State $\mathbold{\hat{y}}=\mathbold{\hat{y}}^{'}$
    \EndIf
\EndFor
\end{algorithmic}
\end{algorithm}

We define our objective function as $f(\mathbold{y}) = \mathbb{E} [\distance(\mathbold{y}, Y)] $. Our objective is to find the global optimal solution $\mathbold{\tilde{y}}$ that minimizes $f(\mathbold{y})$. An optimization algorithm is a procedure that is executed iteratively by comparing various solutions till an optimum or a satisfactory solution is found. We adopt a hill climbing algorithm in this paper, which is an iterative algorithm that attempts to make an incremental change to the current solution every step. The detail of this algorithm is described in Algorithm~\ref{algorithm:1}. Then we can obtain $\mathbold{\hat{y}}$ that is very close to the global optimal solution $\mathbold{\tilde{y}}$:
\begin{equation}
\mathbold{\hat{y}} = \mathop{\argmin}_{\mathbold{y} \in \mathcal{Y}}{\mathbb{E} [\distance(\mathbold{y}, Y)]},
\end{equation}
where $\mathcal{Y}$ is the search space of the optimization procedure. Now for each $x$ in the training set, we can use the above steps to generate a heatmap $\heatmap$, and then use the optimization algorithm to get $\mathbold{\hat{y}}$.

\begin{table*}[tb]
\small
\centering
\begin{tabular}{c|l|ccc}
\toprule
                                & \multicolumn{1}{c|}{Method}                  & minADE               & minFDE               & \cellcolor[HTML]{EFEFEF}Miss Rate            \\ \hline
\multirow{4}{*}{Validation Set} & DESIRE~\cite{cvae}                           & 0.92                 & 1.77                 & \cellcolor[HTML]{EFEFEF}18\%                 \\
                                & MultiPath~\cite{multipath}                   & 0.80                 & 1.68                 & \cellcolor[HTML]{EFEFEF}14\%                 \\
                                & TNT~\cite{tnt}                               & 0.73                 & 1.29                 & \cellcolor[HTML]{EFEFEF}9.3\%                \\
                                & LaneRCNN~\cite{lanercnn}                     & 0.77                 & 1.19                 & \cellcolor[HTML]{EFEFEF}8.2\%                \\ \cline{2-5}
                                & \ourmethod~w/ 100ms optimization & 0.80                 & 1.27                 & \cellcolor[HTML]{EFEFEF}\textbf{7.0\%}                \\
                                & \ourmethod~w/ 100ms optimization (minFDE) & \textbf{0.73}         &  \textbf{1.05} & \cellcolor[HTML]{EFEFEF}9.8\%                \\
                                & \ourmethod~w/ goal set predictor (online)        & 0.82                 & 1.37                 & \cellcolor[HTML]{EFEFEF}\textbf{7.0\%}                \\ \bottomrule

\specialrule{0em}{3pt}{3pt}
\toprule
                                & \multicolumn{1}{c|}{Method}                          & minADE               & minFDE               & \cellcolor[HTML]{EFEFEF}Miss Rate            \\ \hline
\multirow{8}{*}{Leaderboard}    & TNT~\cite{tnt} $8^{th}$                                 & 0.94                 & 1.54                 & \cellcolor[HTML]{EFEFEF}13.3\%               \\
                                & LaneRCNN~\cite{lanercnn} $5^{th}$                       & 0.90                 & 1.45                 & \cellcolor[HTML]{EFEFEF}12.3\%               \\
                                & SenseTime\_AP $4^{th}$                                 & 0.87                 & 1.36                 & \cellcolor[HTML]{EFEFEF}12.0\%               \\
                                & poly $3^{rd}$                                           & \textbf{0.87}                 & 1.47                 & \cellcolor[HTML]{EFEFEF}12.0\%               \\
                                & PRIME~\cite{prime} $2^{nd}$                             & 1.22                 & 1.56                 & \cellcolor[HTML]{EFEFEF}11.5\%               \\
                                & Huawei\_IOV\_France $1^{st}$                            & 0.91                 & 1.36                 & \cellcolor[HTML]{EFEFEF}11.2\%               \\ \cline{2-5}
                                & \ourmethod~w/ 100ms optimization        & 0.94                 & 1.49                 & \cellcolor[HTML]{EFEFEF}\textbf{10.5\%}               \\
                                & \ourmethod~w/ 100ms optimization (minFDE)& 0.88                 & \textbf{1.28}                 & \cellcolor[HTML]{EFEFEF}12.6\%               \\
                                & \ourmethod~w/ goal set predictor (online)               & 0.93                 & 1.45                 & \cellcolor[HTML]{EFEFEF}10.7\%                \\ \bottomrule
\end{tabular}
\caption{\label{table:rank} Model performance on the Argoverse validation set and leaderboard (as of March 16, 2021). Miss Rate is the official ranking metric.}
\end{table*}

\paragraph{Goal set predictor (online).}
Set predictor was introduced by DETR~\cite{detr}, which views object detection as a set prediction problem and designs a loss based on the Hungarian matching.
In this multi-future prediction problem, we treat it as a set prediction problem as well and use the output of the offline model as pseudo-labels to train the goal set predictor of the online model.
Instead of performing Hungarian matching between the set of predicted goals and pseudo-labels, we perform offline optimization during training, using each optimized pseudo-label to supervise its corresponding predicted goal.

Let us denote $\Dot{\mathbold{y}}=\{\Dot{y}_i\}_{i=1}^{K}$ as the set of $K$ predicted goals generated by the goal set predictor at the current training step.
We use the above optimization algorithm to generate the pseudo-labels $\mathbold{\hat{y}}$ for this training step. The initial goal set of the optimization algorithm is set to the predicted goal set $\Dot{\mathbold{y}}$. The optimization algorithm only searches for the neighbors of $\Dot{\mathbold{y}}$ instead of searching for the optimal solution. Specifically, we run the random perturbation for $L (L=100)$ times to get $L$ goal sets. The pseudo-labels $\mathbold{\hat{y}}$ for the goal set predictor at the current training step is the goal set with the lowest expected error.

The loss term is the offset between the predicted goal set $\Dot{\mathbold{y}}$ and the pseudo-labels $\mathbold{\hat{y}}$:
\begin{equation}
\mathcal{L}_{\rm set}(\Dot{\mathbold{y}}, \mathbold{\hat{y}}) = \sum\limits ^{K}_{i=1} \mathcal{L}_{\rm reg} (\Dot{y}_i, \hat{y}_i),
\end{equation}
where $\mathcal{L}_{\rm reg}$ is the standard $\ell_1$ loss between two goals.

Since the probability distribution indicated by the heatmap is diverse, it is difficult for a single regressor to handle. The goal set predictor has multiple heads to predict $N$ goal sets simultaneously. Specifically, every head will predict $2K+1$ values, including the 2D coordinates of $K$ goals and the confidence of this head.
Every head is composed of a heatmap encoder and a decoder. The heatmap encoder is a one-layer self-attention followed by a max-pooling, and the decoder is a two-layer MLP that outputs $2K+1$ values. The parameters of the heatmap encoders of all heads are shared to reduce computation.

During training, the optimization algorithm only generates pseudo-labels for the head with the lowest expected error and the goal set predictor only performs regression on this head. To predict the confidences of the multiple heads, we use a binary cross-entropy loss:
\begin{equation}
\mathcal{L}_{\rm head} = \mathcal{L}_{\rm CE}(\mathboldgreek{\mu},\mathboldgreek{\nu}),
\end{equation}
where $\mathboldgreek{\mu}$ is the predicted confidence of the heads, and $\mathboldgreek{\nu}$ is the confidence label. $\nu_i = 1$ for the head with the lowest expected error and $\nu_i = 0$ for other heads.
During inference, we take the head with the highest confidence as the output of the goal set predictor.

\subsection{Trajectory completion}
Similar to TNT, the last step is to complete each trajectory conditioned on the predicted goals. We first calculate the feature of each goal similar to the above dense goal encoding, then pass it to the decoder that is a 2-layer MLP. The output of the decoder is the whole trajectory $[\hat{s}_1, \hat{s}_2, \dots, \hat{s}_T]$.

We only have one ground truth trajectory, so we apply a teacher forcing technique~\cite{williams1989learning} by feeding the ground truth goal during training.
The loss term is the offset between the predicted trajectory $\mathbold{\hat{s}}$ and the ground truth trajectory $\mathbold{s}$:
\begin{equation}
\mathcal{L}_{\rm completion} = \sum\limits ^{T}_{t=1} \mathcal{L}_{\rm reg} (\hat{s}_t, s_t),
\end{equation}
where $\mathcal{L}_{\rm reg}$ is the smooth $\ell_1$ loss between two points. During inference, this trajectory completion module is used to generate $K$ trajectories of the $K$ goals simultaneously.

\subsection{Learning}
\label{sec:learning}
The training procedure of our method has two stages. In the first stage, we use the ground truth trajectories to train all the modules except for the goal set predictor:
\begin{equation}
\mathcal{L}_{\rm S1} = \mathcal{L}_{\rm lane} + \mathcal{L}_{\rm goal} + \mathcal{L}_{\rm completion}.
\end{equation}
In the second stage, we train the goal set predictor on the training set, which is supervised by the pseudo-labels generated by the offline model (encoding + optimization algorithm):
\begin{equation}
\mathcal{L}_{\rm S2} = \mathcal{L}_{\rm head} + \mathcal{L}_{\rm set}.
\end{equation}

\begin{table*}[tbp]
\centering
\small
\begin{tabular}{c|cc|c|ccc|ccc}
\toprule
\multicolumn{4}{r|}{Optimization Objective}                                                                                                                                                    & \multicolumn{3}{c|}{minFDE}                                                                                 & \multicolumn{3}{c}{MR}                                                                                     \\ \hline
                                                                                   & \multicolumn{2}{c|}{Probability estimation}  &                                                                            &                          & \cellcolor[HTML]{EFEFEF}                                  &                      &                          &                          & \cellcolor[HTML]{EFEFEF}                              \\ \cline{2-3}
 \multirow{-2}{*}{\begin{tabular}[c]{@{}c@{}}     End-to-end        \end{tabular}} & Sparse        & Dense                        & \multirow{-2}{*}{\begin{tabular}[c]{@{}c@{}}Goal   selection\end{tabular}} & \multirow{-2}{*}{minADE} & \multirow{-2}{*}{\cellcolor[HTML]{EFEFEF}\textbf{minFDE}} & \multirow{-2}{*}{MR} & \multirow{-2}{*}{minADE} & \multirow{-2}{*}{minFDE} & \multirow{-2}{*}{\cellcolor[HTML]{EFEFEF}\textbf{MR}} \\ \hline
 \checkmark                                                                        &               &                              & variety loss                                                               & 0.78                      & \cellcolor[HTML]{EFEFEF}1.25                             & 13.3\%               & 0.78                     & 1.25                     & \cellcolor[HTML]{EFEFEF}13.3\%                           \\ \hline
                                                                                   & \checkmark    &                              & NMS                                                                        & 0.81                     & \cellcolor[HTML]{EFEFEF}1.33                     & 10.1\%               & 0.82                     & 1.35                     & \cellcolor[HTML]{EFEFEF}9.5\%                \\ \hline
                                                                                   &               & \checkmark                   & NMS                                                                        & 0.79                     & \cellcolor[HTML]{EFEFEF}1.26                     & 8.6\%                & 0.79                     & 1.28                     & \cellcolor[HTML]{EFEFEF}8.2\%                \\ \hline
                                                                                   &               & \checkmark                   & optimization                                                               & 0.73                     & \cellcolor[HTML]{EFEFEF}\textbf{1.05}                     & 9.8\%                & 0.80                     & 1.28                     & \cellcolor[HTML]{EFEFEF}\textbf{7.0\%}                \\ \hline
 \checkmark                                                                        &               & \checkmark                   & goal set predictor                                                         & 0.75                     & \cellcolor[HTML]{EFEFEF}\textbf{1.05}                & 9.7\%                & 0.82                     & 1.37                     & \cellcolor[HTML]{EFEFEF}\textbf{7.0\%} \\ \bottomrule
\end{tabular}
\caption{\label{table:ablation} Ablation studies on the main components in our method: dense probability estimation, goal selection, \etc. We also tested the effectiveness of our method under different optimization objectives, minFDE and MR.}
\end{table*}

\section{Experiments}

\subsection{Datasets}
\paragraph{Argoverse forecasting dataset.}
Argoverse forecasting dataset~\cite{argoverse} is a dataset with agent trajectories and high-definition maps. Given the trajectory of the target vehicle in the past two seconds, which are sampled at 10Hz, we need to predict the future trajectory in the next 3 seconds. There are 333K real-world driving sequences that are at intersections or in dense traffic, and each sequence contains one target vehicle for prediction.  The training, validation, and test sets contain 205942, 39472, and 78143 sequences, respectively.

\paragraph{Waymo open motion dataset.}
Waymo open motion dataset~\cite{waymo_open} is by far the most diverse interactive motion dataset. It contains more than 570 hours of unique data over 1750km of roadways with over 100,000 scenes, each 20 seconds long. There are three types of agents in the dataset, namely vehicles, pedestrians and cyclists. Given a 1-second history trajectory of the target agent, an 8-second future trajectory needs to be predicted.

\paragraph{Metrics.}
We follow the Argoverse benchmark and use minimum average displacement error (minADE), minimum final displacement error (minFDE), and miss rate (MR).
Each trajectory is represented by a sequence of points over time. ADE is the average displacement between each point of the predicted trajectory and its corresponding ground-truth point. minADE is the minimum ADE of the predicted $K$ trajectories, and minFDE is the minimum displacement between $K$ final positions and the ground truth final position.
Miss rate is the ratio of scenarios where none of the predicted trajectories are within 2.0 meters of ground truth according to the final displacement error.

\subsection{Implementation Details}
\label{section:detail}

\begin{figure*}[tbh]
\centering
\setlength{\fboxsep}{0pt}
\begin{minipage}[b]{0.23\linewidth}
\fbox{\includegraphics[width=42mm, height=42mm]{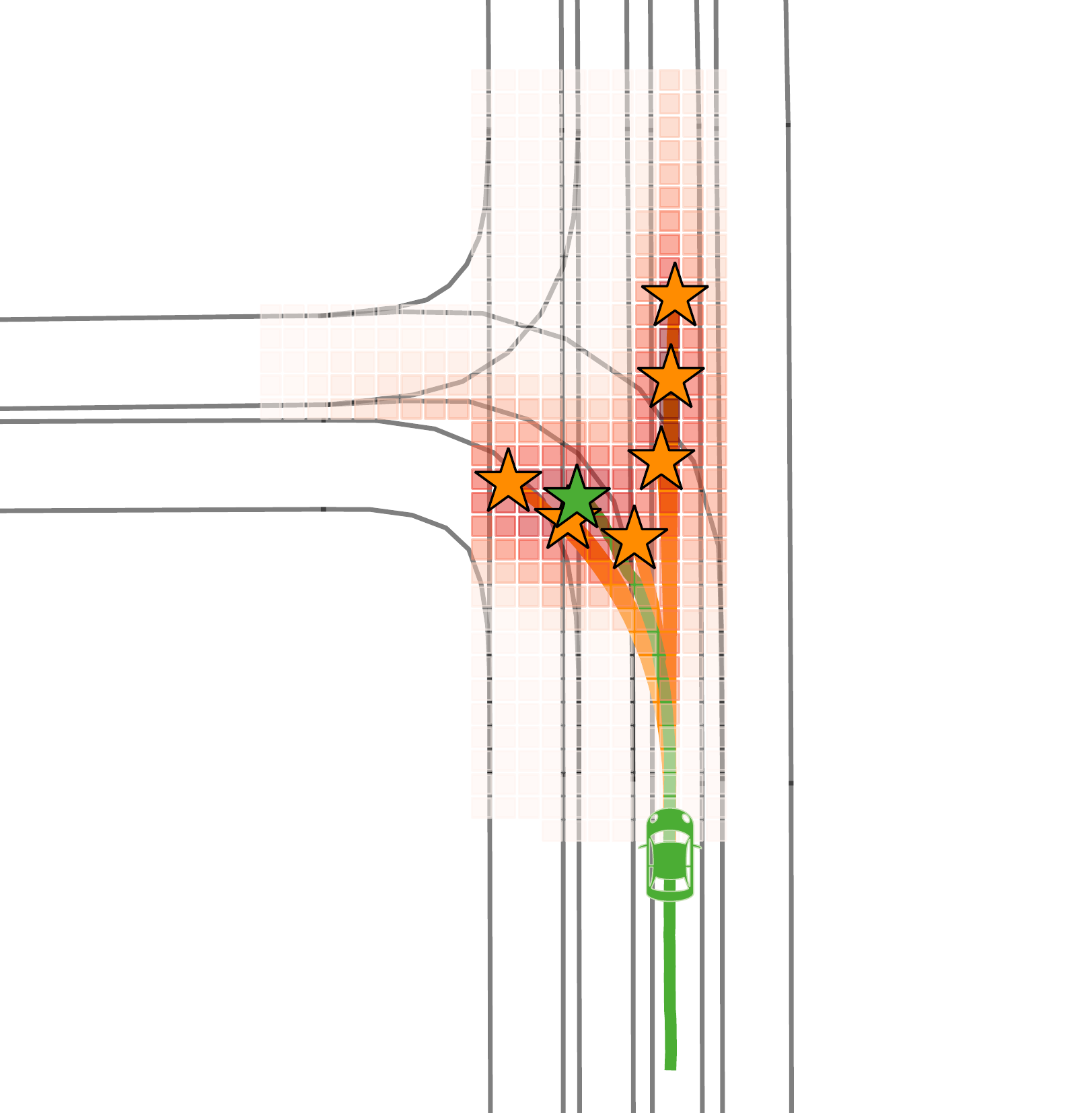}}
\end{minipage}
\hspace{4pt}
\begin{minipage}[b]{0.23\linewidth}
\fbox{\includegraphics[width=42mm, height=42mm]{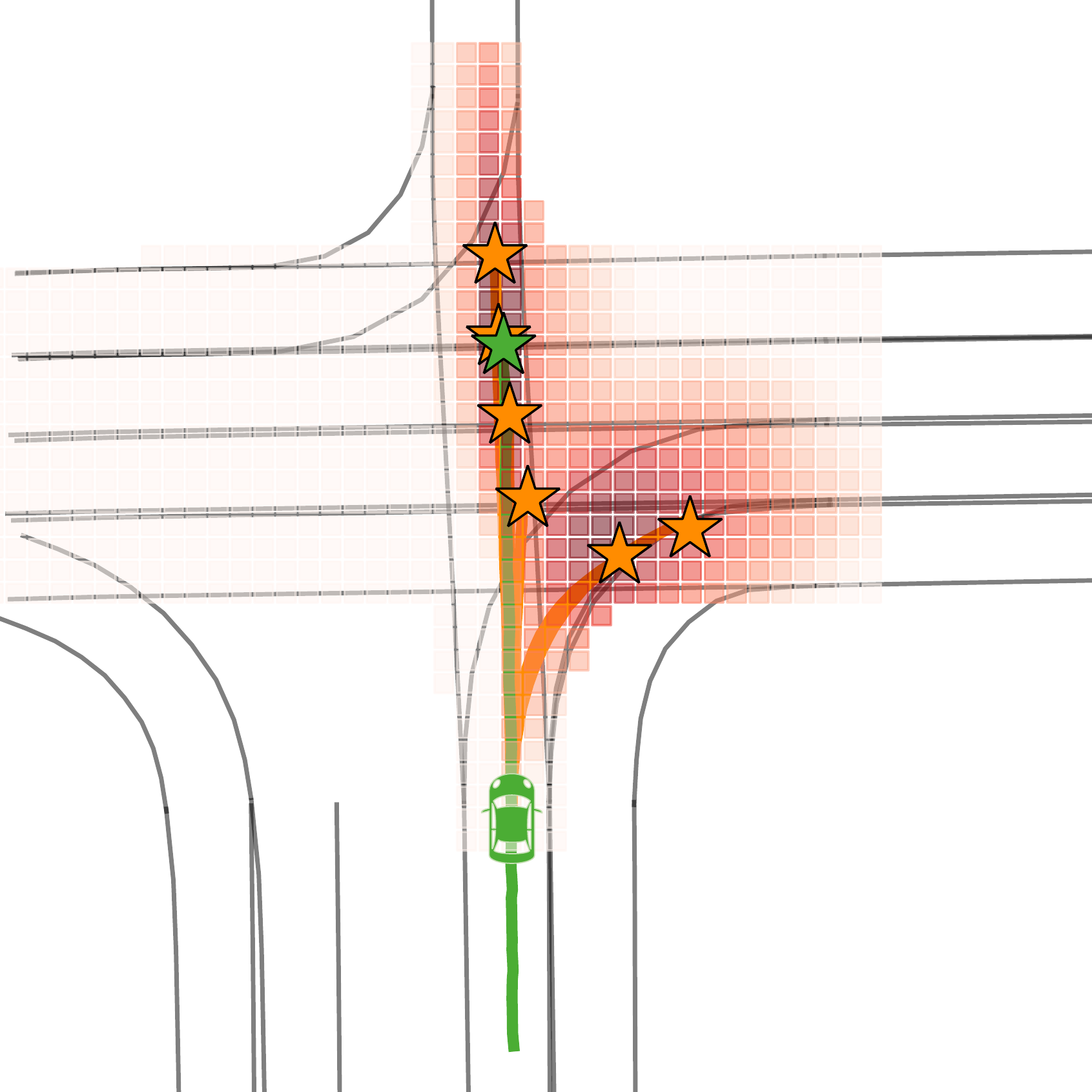}}
\end{minipage}
\hspace{4pt}
\begin{minipage}[b]{0.23\linewidth}
\fbox{\includegraphics[width=42mm, height=42mm]{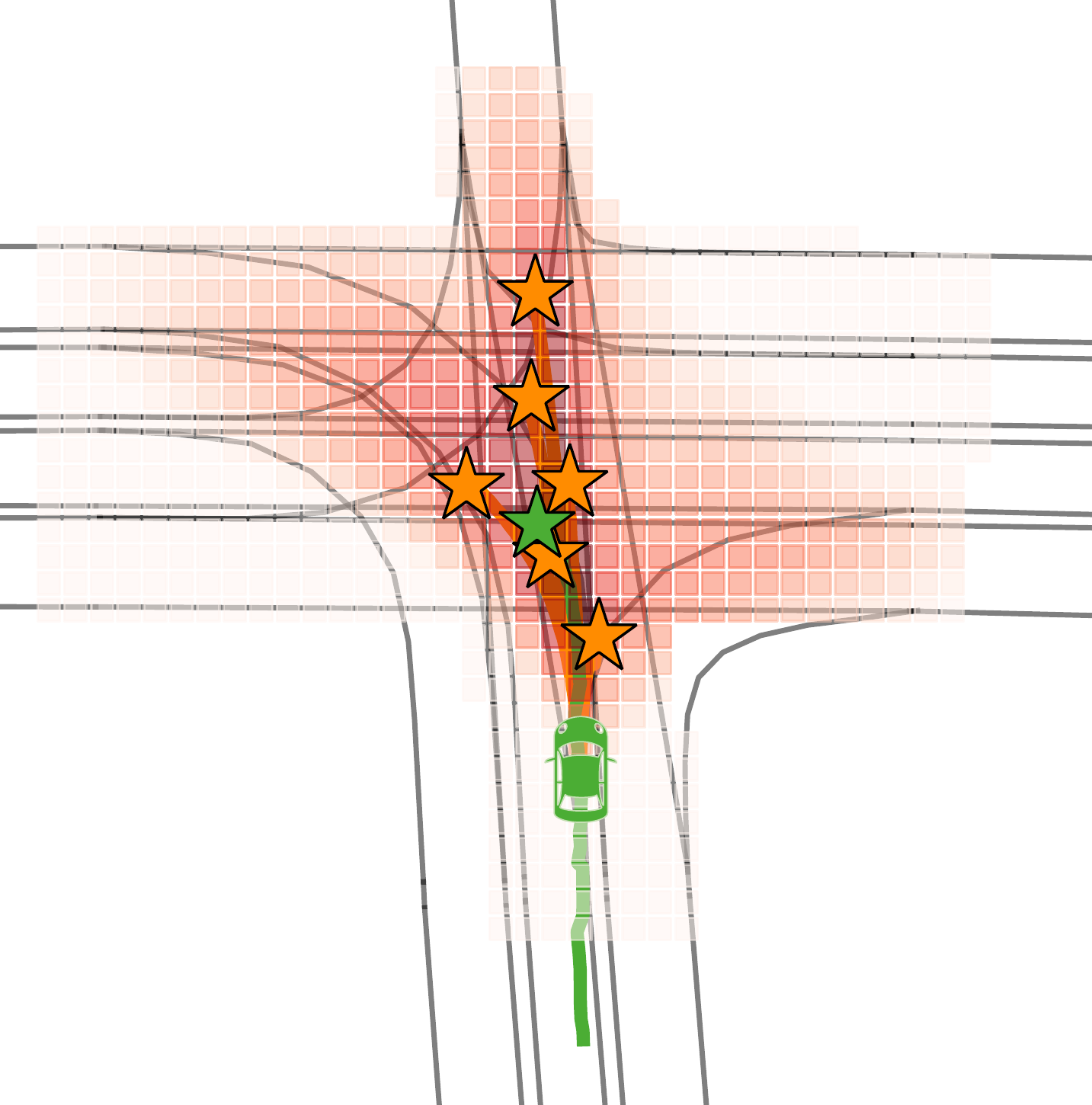}}
\end{minipage}
\hspace{4pt}
\begin{minipage}[b]{0.23\linewidth}
\fbox{\includegraphics[width=42mm, height=42mm]{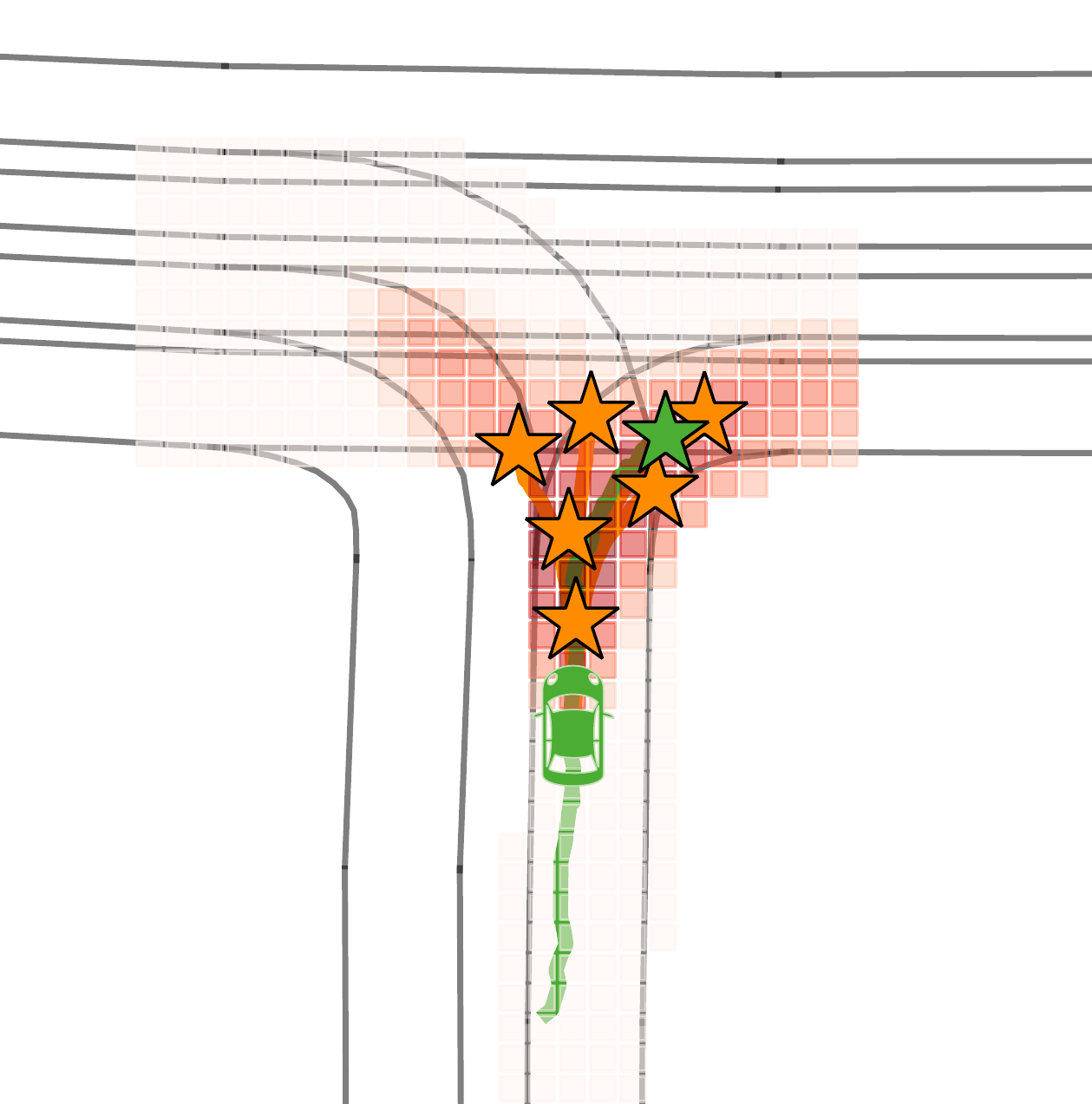}}
\end{minipage}
\caption{\label{visual}Qualitative results of DenseTNT (online). Dense predicted heatmaps are shown in red, predicted goal sets and corresponding trajectories are shown in orange, ground truth trajectories are shown in green.}
\end{figure*}

\paragraph{Goal candidate sampling.}
We first sample the lanes within 50m (Manhattan distance) from the target vehicle. Then we sample goal candidates which are densely distributed on these lanes. Therefore, the number of the sampled goal candidates depends on the lanes around the target vehicle. For the lanes represented by the lane centerlines, the goal candidates within 3m from the centerlines are sampled, while for the lanes represented by lane boundaries, the goal candidates within the boundaries are sampled. The distance between two adjacent goals, \ie, the sampling density, is set to 1m.

\paragraph{Training details.}
Our model is trained on the training set with a batch size of 64. In the first stage, we train all the modules except for the goal set predictor for 16 epochs, and the learning rate with an initial value of 0.001 decays to 30\% every 5 epochs. In the second stage, we train the goal set predictor for 6 epochs, and the learning rate with an initial value of 0.001 decays to 30\% every epoch. The hidden size of the feature vectors is set to 128. The head number of our goal set predictor is 12. No data augmentation is used. 

\subsection{Results on benchmarks}
\paragraph{Argoverse motion forecasting benchmark.}
We evaluate \ourmethod~on the Argoverse validation set, and report results in Table~\ref{table:rank}. As can be seen, \ourmethod~greatly outperforms popular models in the literature.
It is also worth noticing that our online model (\ourmethod~w/ goal set predictor), though trained from the pseudo-labels provided by the offline model (\ourmethod~w/ optimization), achieves comparable results as the offline model.
We further compare \ourmethod~with the top performers on the Argoverse leaderboard in  Table~\ref{table:rank}. Since the details of the 1st, 3rd and 4th methods were undisclosed, we could not compare them qualitatively. Our method can generate the trajectories in an end-to-end manner during real-time usage, in contrast with PRIME (2nd) and LaneRCNN (5th) that use NMS to perform post-processing.
We achieve superior performance on the official ranking metric MR, which verifies the effectiveness of our method. For another popular metric minFDE, we can also achieve state-of-the-art performance by using it as the optimization objective.

Figure~\ref{visual} shows the qualitative results generated by our online model. The probability distribution of the goals in some cases is quite multimodal, making it difficult for NMS to handle in the post-processing stage. Our model makes diverse trajectory predictions with high coverage of the heatmaps. 

\begin{table}[tb]
\centering
\small
\begin{tabular}{lcccc}
\toprule
Method                                & mADE & mFDE       & MR  &\cellcolor[HTML]{EFEFEF} mAP    \\ \hline
DenseTNT $1^{st}$ (Ours)              & 1.0387        & 1.5514    & 0.1779      &\cellcolor[HTML]{EFEFEF} \bf 0.3281 \\
TVN  $2^{nd}$                         & 0.7558 & 1.5859       & 0.2032     &\cellcolor[HTML]{EFEFEF} 0.3168 \\
Star Platinum $3^{rd}$                & 0.8102 & 1.7605     & 0.2341     &\cellcolor[HTML]{EFEFEF} 0.2806 \\
SceneTransformer\cite{ngiam2021scene} & 0.6117 & 1.2116 & 0.1564  &\cellcolor[HTML]{EFEFEF} 0.2788 \\
ReCoAt                                & 0.7703    & 1.6668    & 0.2437     &\cellcolor[HTML]{EFEFEF} 0.2711 \\ \bottomrule
\end{tabular}
\caption{\label{table:waymo} Top 5 entries of the 2021 Waymo Open Dataset Motion Prediction Challenge. mAP is the official ranking metric.}
\end{table}

\paragraph{Waymo Open Dataset Motion Prediction Challenge.}
We developed a variant of DenseTNT for the 2021 Waymo Open Dataset Motion Prediction Challenge, and got the $1^{st}$ place.
The challenge leaderboard is shown in Table~\ref{table:waymo}. Details of this variant are discussed in our technical report\footnote{Available on \url{https://waymo.com/open/challenges}.}.

\subsection{Ablation Study}
\paragraph{Model architecture.}
We conduct ablation studies on the main components of our model. These components are the dense probability estimation, the optimization algorithm for generating pseudo-labels, and the goal set predictor. There are different metrics to measure the performance of generating the most likely trajectories. We tested the effectiveness of our method under different optimization objectives, as shown in Table~\ref{table:ablation}.

Each component plays an important role in our method. The dense probability estimation performs much better than the sparse probability estimation, since the dense probability estimation provides more fine-grained local information. Moreover, the sparse probability estimation can only be combined with NMS which is a heuristic rule-based algorithm. The hyper-parameter of NMS is the threshold of removing neighboring points, \ie, two points with a distance less than the threshold are regarded as the same point. For a fair comparison, we show the best result of NMS under different metrics. The result of the online model is almost the same as the offline model, proving the effectiveness of the goal set predictor. Variety loss is a traditional end-to-end trajectory prediction method, which generates a fixed number of trajectories but only performs regression on the closest one during training. Our end-to-end method outperforms it by a large margin.

\paragraph{Goal density.}
To denote the probability distribution of the final position, we densely sample goal candidates on the lanes. The sampling density of the goals has an impact on our method's performance, and we show it in Table~\ref{table:density}. It indicates that a higher density leads to better performance until the saturation point is reached.

\begin{table}[htbp]
\centering
\small
\begin{tabular}{c|c|c}
\toprule
Sampling density & minFDE & MR \\ \hline
3.0m             &   1.42   & 12.5\%\\
2.0m             &   1.34   & 9.0\%\\
1.0m             &   1.27   & 7.0\%\\
0.5m             &   1.27   & 7.0\% \\ \bottomrule
\end{tabular}
\caption{\label{table:density} Comparison of different goal sampling densities on the Argoverse validation set.}
\end{table}

\begin{table}[htb]
\centering
\small
\begin{tabular}{c|c|c}
\toprule
Optimization time& minFDE       & MR  \\ \hline
20ms             &   1.29      & 7.6\% \\
50ms             &   1.28      & 7.2\% \\
100ms             &    1.27      & 7.0\% \\
200ms             &    1.27      & 6.9\% \\
500ms             &    1.27      & 6.9\% \\ \bottomrule
\end{tabular}
\caption{\label{table:optimization}Performance under different optimization time per sample on the Argoverse validation set.}
\end{table}

\paragraph{Optimization.}
Given a heatmap indicating the probability distribution of the vehicle's final position, the optimization algorithm is used to find the global optimal solution. The maximum running time of the optimization algorithm per instance has an impact on the performance. Table~\ref{table:optimization} shows the optimization performance over time. The performance increases drastically before $t=100{\rm ms}$ and remains almost unchanged after $t=200{\rm ms}$.


\section{Conclusion}
In this paper, we propose an anchor-free and end-to-end trajectory prediction model, named DenseTNT, that directly outputs a set of trajectories from dense goal candidates. In addition, we introduce an optimization-based offline model to provide multi-future pseudo-labels to train the online model. DenseTNT not only runs online, but also has similar performance as the offline model, demonstrating the effectiveness of the goal set predictor design and our training paradigm. Comprehensive experiments show that DenseTNT achieves state-of-the-art performance, ranking $1^{st}$ on the Argoverse motion forecasting benchmark and being the $1^{st}$ place winner of the 2021 Waymo Open Dataset Motion Prediction Challenge.

\paragraph{Acknowledgments.}
We would like to thank Dr. Jiyang Gao and Chenzhuang Du for helpful discussions on the manuscript.

{\small
\bibliographystyle{ieee_fullname}
\bibliography{egbib}
}

\clearpage
\appendix

\section{Offline Optimization}
To boost the training of DenseTNT, we devise an offline model which is composed of a context encoding module and an optimization algorithm.
There are different metrics to measure the performance of multi-trajectory prediction methods. For a comprehensive evaluation,
we tested the effectiveness of the optimization algorithm under different combinations of optimization objectives, as shown in Table~\ref{appendix:rank}.

\section{Implementation Details}
\paragraph{Agent and map encoding.}
To normalize the map, we take the last position of the target vehicle as the origin and the direction of the target vehicle as the $y$-axis. Following VectorNet~\cite{vectornet}, the lanes and the agents are converted into sequences of vectors. Each vector contains the start point, the endpoint, and the attributes of its corresponding lane or agent. A vector that belongs to a lane also contains its index in this lane, and a vector that belongs to an agent contains the timestamps of its start point and end point. After the sparse context encoding, we obtain the features of the lanes and the agents.

\paragraph{Optimization algorithm.}
The optimization algorithm aims to find a goal set that minimizes the expected error. It is implemented by a statically-typed language to achieve the fastest speed and search for hundreds of goal sets in 100ms. We run the optimization algorithm on 8 CPUs in parallel with different initializations and pick the best result. The main cost is on the calculation of the expected error of each searched goal set.

The probability distribution of the final position is indicated by heatmap goals $\mathcal{C}=\{c_1,c_2,...,c_m\}$ and their corresponding probabilities $h(c_i)$. When calculating the expected error of a given goal set, we only consider $c_i$ which satisfies $h(c_i) \geq 10^{-3}$.

\vfill
\noindent
\begin{minipage}{1.0\textwidth}
\strut\newline
\centering
\begin{tabular}{c|l|cc|ccc}
\toprule
                                & \multicolumn{3}{c|}{Method}                                               & minADE               & minFDE               & \cellcolor[HTML]{EFEFEF}Miss Rate            \\ \hline
\multirow{8}{*}{Validation Set} & \multicolumn{3}{l|}{DESIRE~\cite{cvae}           }                        & 0.92                 & 1.77                 & \cellcolor[HTML]{EFEFEF}18\%                 \\
                                & \multicolumn{3}{l|}{MultiPath~\cite{multipath}   }                        & 0.80                 & 1.68                 & \cellcolor[HTML]{EFEFEF}14\%                 \\
                                & \multicolumn{3}{l|}{TNT~\cite{tnt}               }                        & 0.73                 & 1.29                 & \cellcolor[HTML]{EFEFEF}9.3\%                \\
                                & \multicolumn{3}{l|}{LaneRCNN~\cite{lanercnn}     }                        & 0.77                 & 1.19                 & \cellcolor[HTML]{EFEFEF}8.2\%                \\ \cline{2-7}
                                & \multirow{6}{*}{DenseTNT w/ optimization (100ms)} & minFDE & Miss Rate           &                      &                      & \cellcolor[HTML]{EFEFEF}          \\ \cline{3-7}
                                &                                                   &\ \ \ \ 0\%    & 100\%        & 0.80                    & 1.27                    & \cellcolor[HTML]{EFEFEF}\textbf{7.0}\%         \\
                                &                                                   &\ \ 30\%       & \ \  70\%    & 0.74                    & 1.12                 & \cellcolor[HTML]{EFEFEF}7.3\%    \\
                                &                                                   &\ \ 50\%       & \ \  50\%    & 0.74                    & 1.09                 & \cellcolor[HTML]{EFEFEF}7.5\%    \\
                                &                                                   &\ \ 70\%       & \ \  30\%    & 0.73                   & 1.08                & \cellcolor[HTML]{EFEFEF}8.0\%    \\
                                &                                                   &100\%          &   \ \ \ \ 0\%& \textbf{0.73}          & \textbf{1.05}                & \cellcolor[HTML]{EFEFEF}9.8\%    \\ \bottomrule

\end{tabular}

\captionof{table}{Performance of the optimization algorithm under different combinations of optimization objectives.}
\label{appendix:rank}
\end{minipage}

Since the sample density is 1m, each heatmap goal $c_i$ represents a space of $1m \times 1m$. To obtain a more precise expected error, we divide each heatmap goal to 9 heatmap goals with a probability of $\frac{1}{9}h(c_i)$, and each of them represents a space of $\frac{1}{3}m \times \frac{1}{3}m$.

\paragraph{Goal set predictor.}
The goal set predictor aims to learn a mapping from the heatmap to the goal set. We only encode heatmap goals which satisfies $h(c_i) \geq 10^{-5}$. First, we normalize both the 2D coordinates of heatmap goals and the pseudo labels by taking the heatmap goal with the highest probability as the origin. Then, a 2-layer MLP is used to encode the heatmap goals, of which input is the 2D coordinates of each goal and its corresponding log probability. The features of heatmap goals are passed to the predictor heads. A softmax function is employed to normalize the predicted confidence of all heads. The head number of the goal set predictor is set to 12.


\section{Qualitative Results}
\begin{figure*}[tbh]
\centering
\setlength{\fboxsep}{0pt}

\begin{minipage}[b]{0.23\linewidth}
\fbox{\includegraphics[width=42mm, height=42mm]{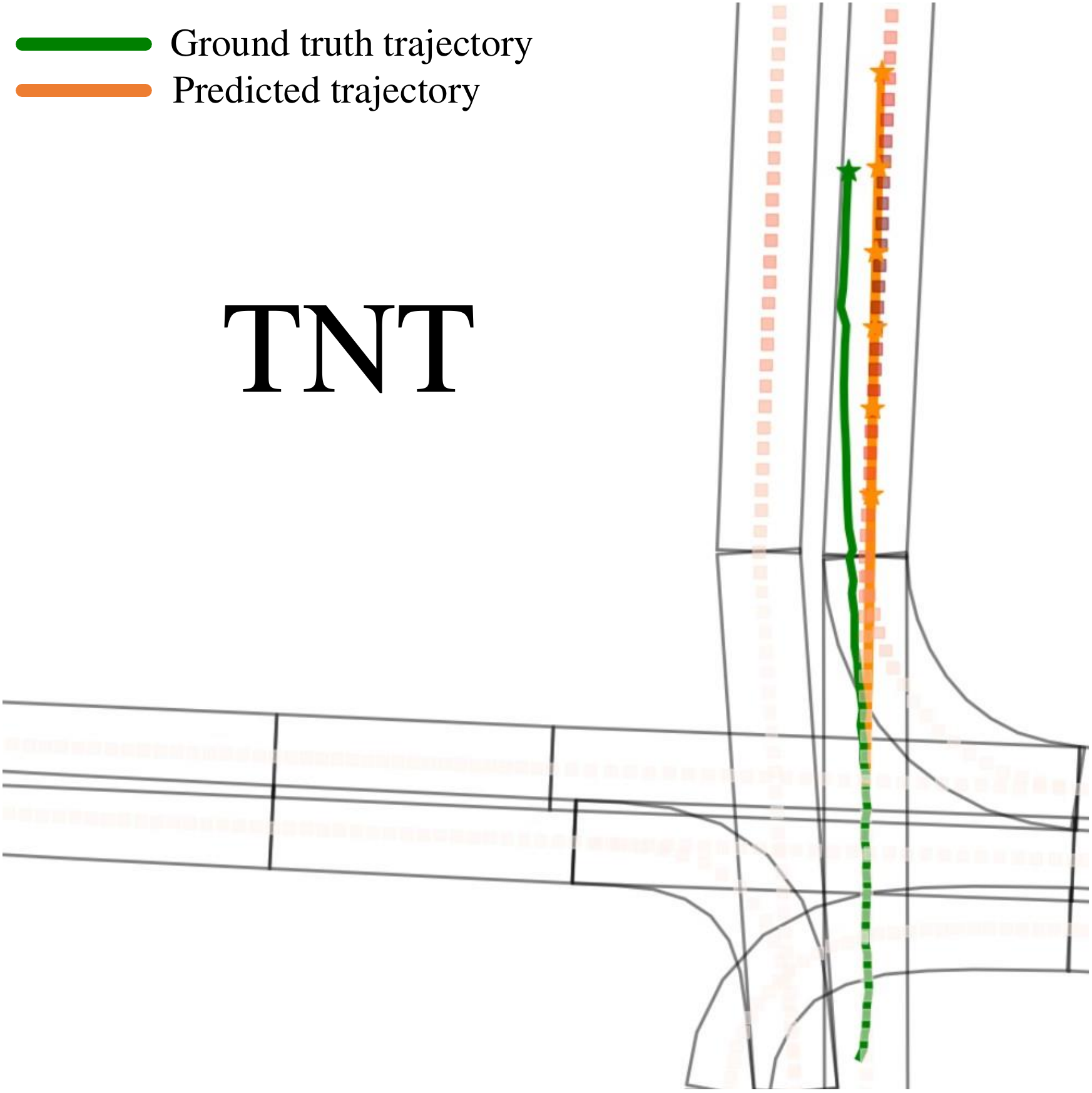}}\vspace{6pt}
\fbox{\includegraphics[width=42mm, height=42mm]{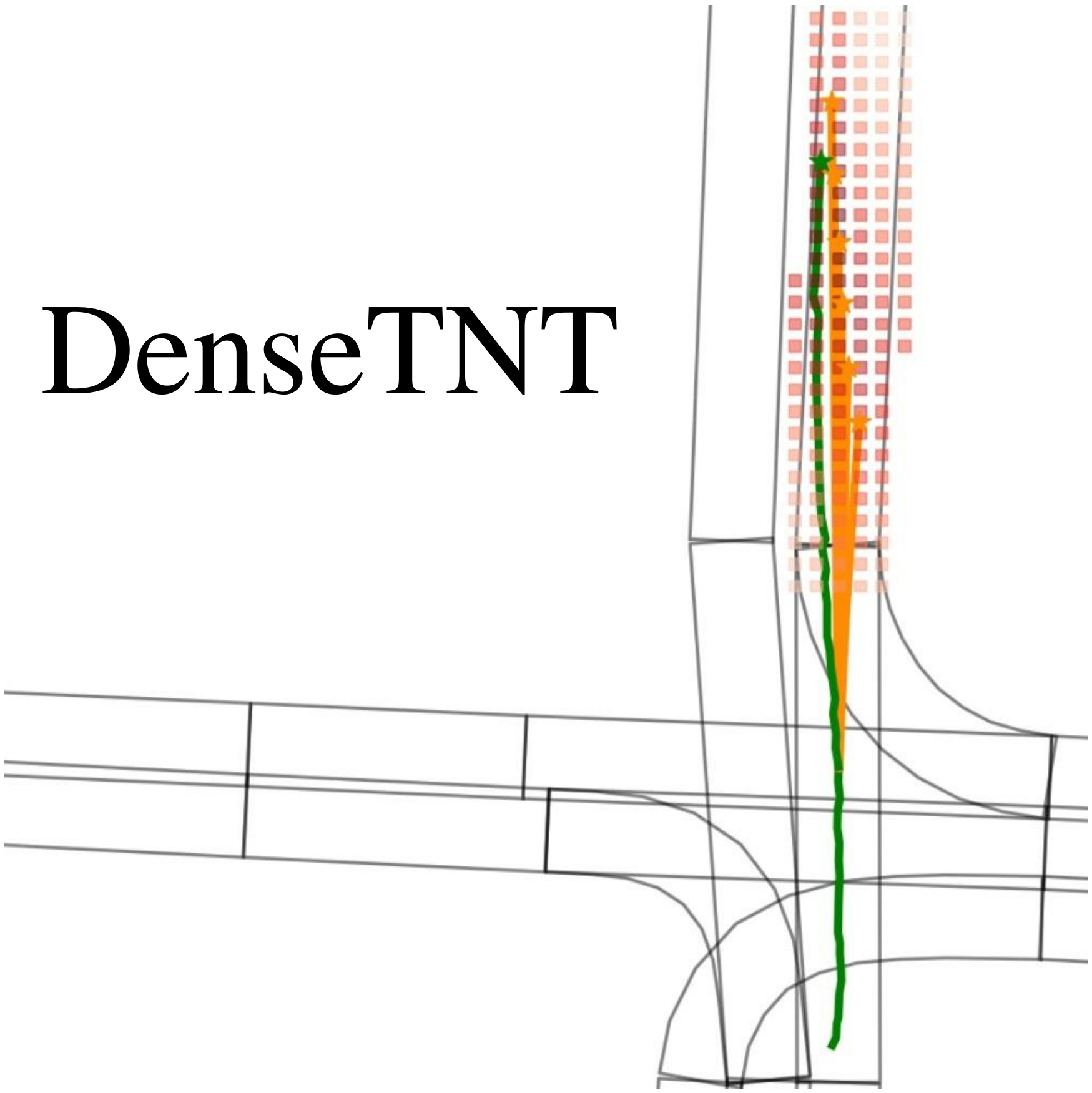}}
\end{minipage}
\hspace{4pt}
\begin{minipage}[b]{0.23\linewidth}
\fbox{\includegraphics[width=42mm, height=42mm]{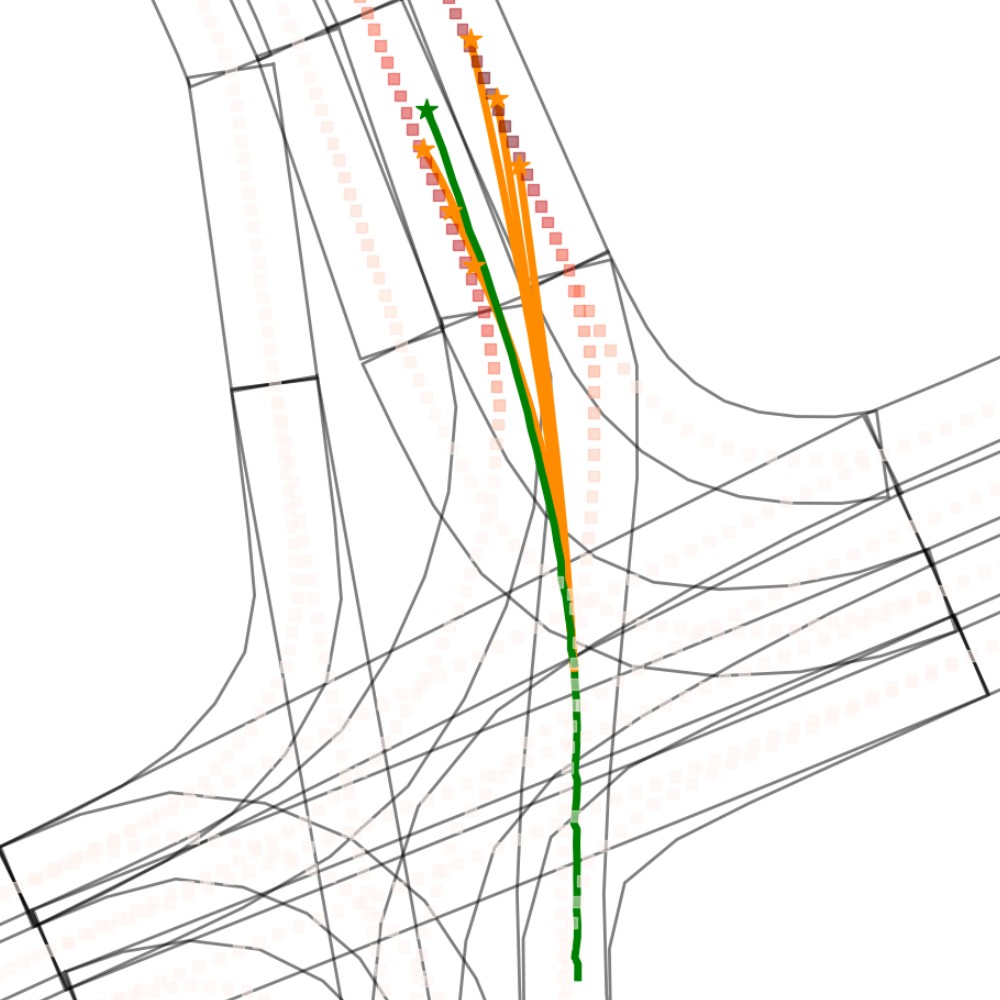}}\vspace{6pt}
\fbox{\includegraphics[width=42mm, height=42mm]{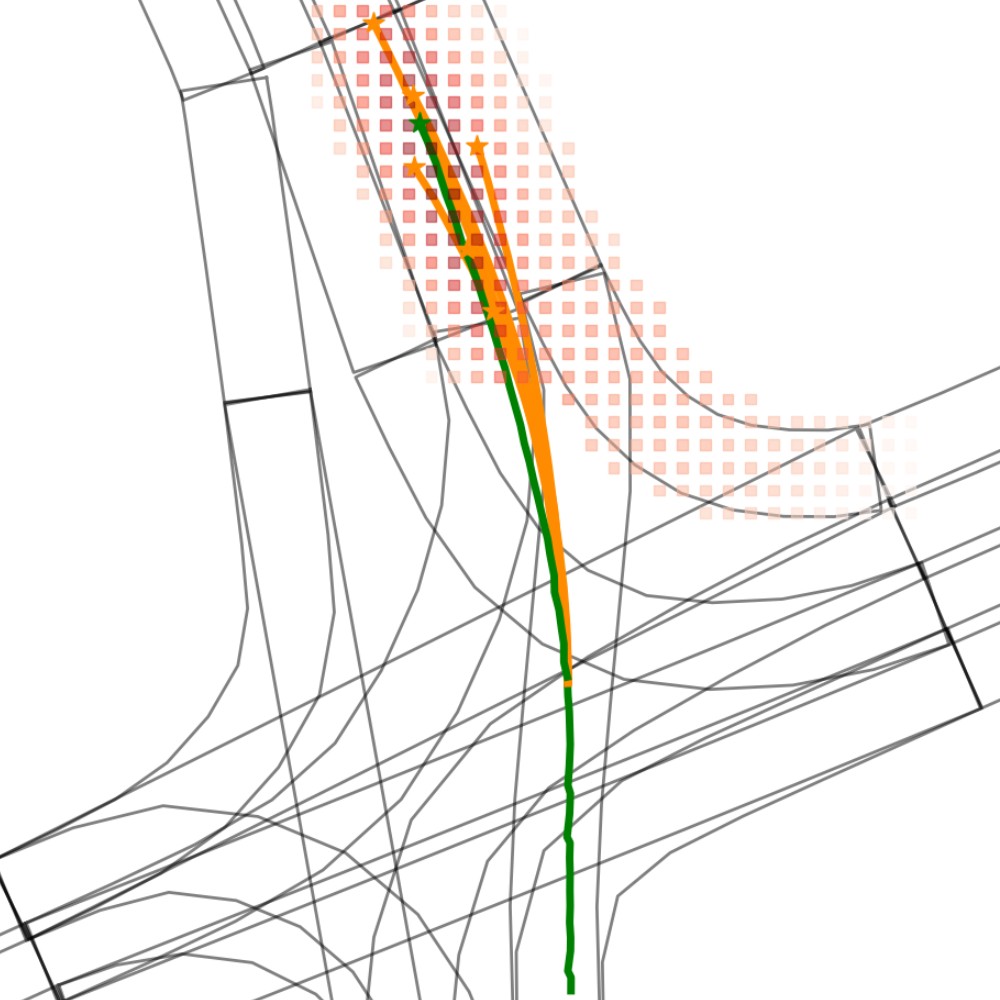}}
\end{minipage}
\hspace{4pt}
\begin{minipage}[b]{0.23\linewidth}
\fbox{\includegraphics[width=42mm, height=42mm]{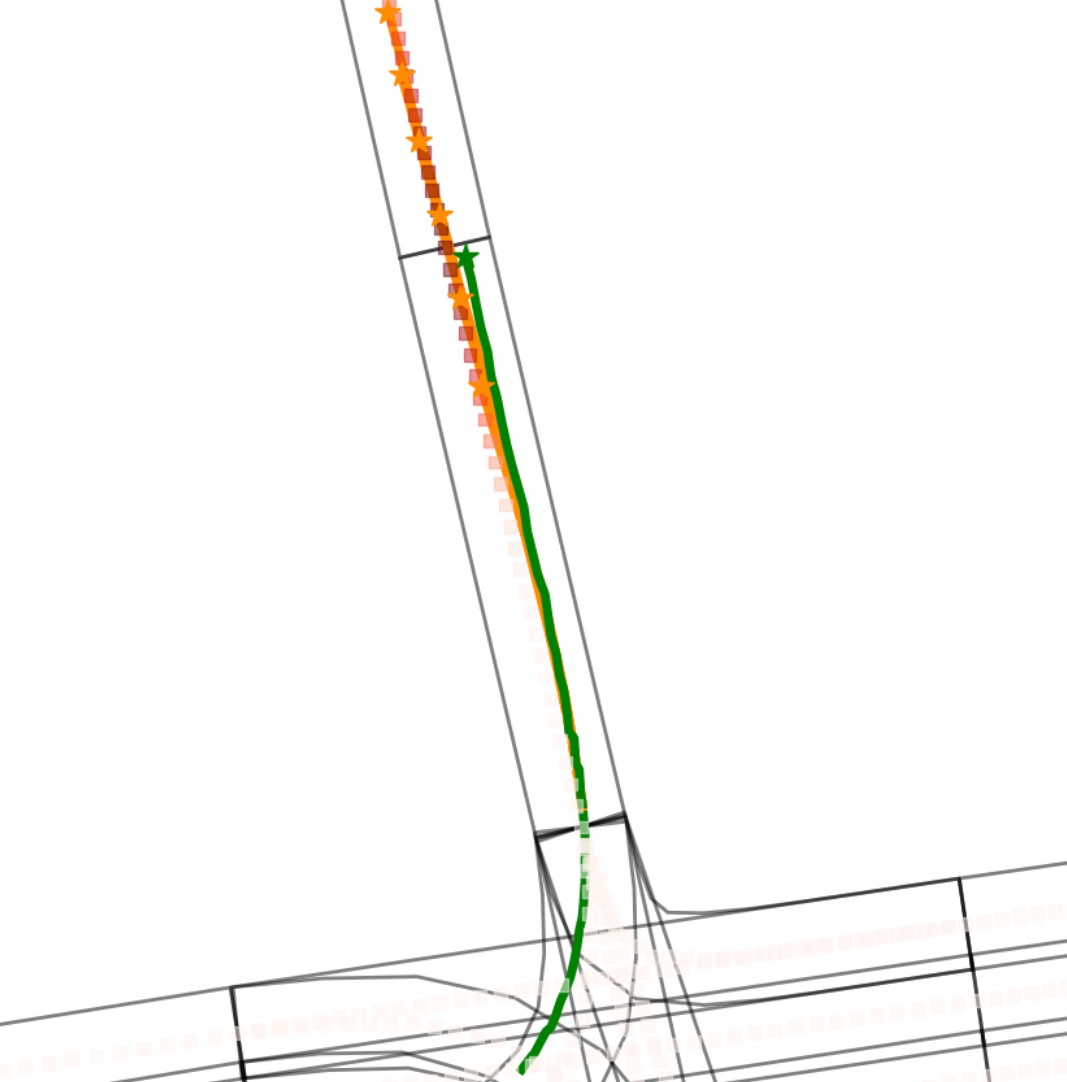}}\vspace{6pt}
\fbox{\includegraphics[width=42mm, height=42mm]{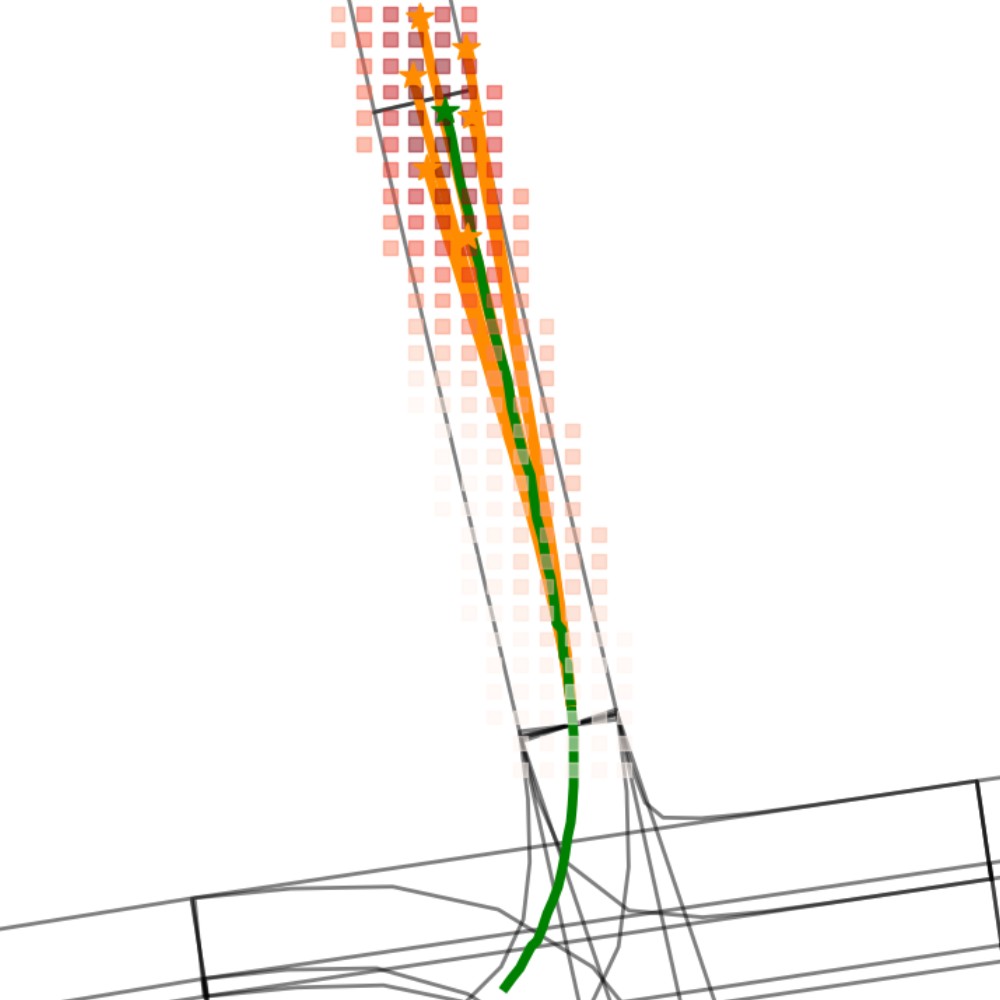}}
\end{minipage}
\hspace{4pt}
\begin{minipage}[b]{0.23\linewidth}
\fbox{\includegraphics[width=42mm, height=42mm]{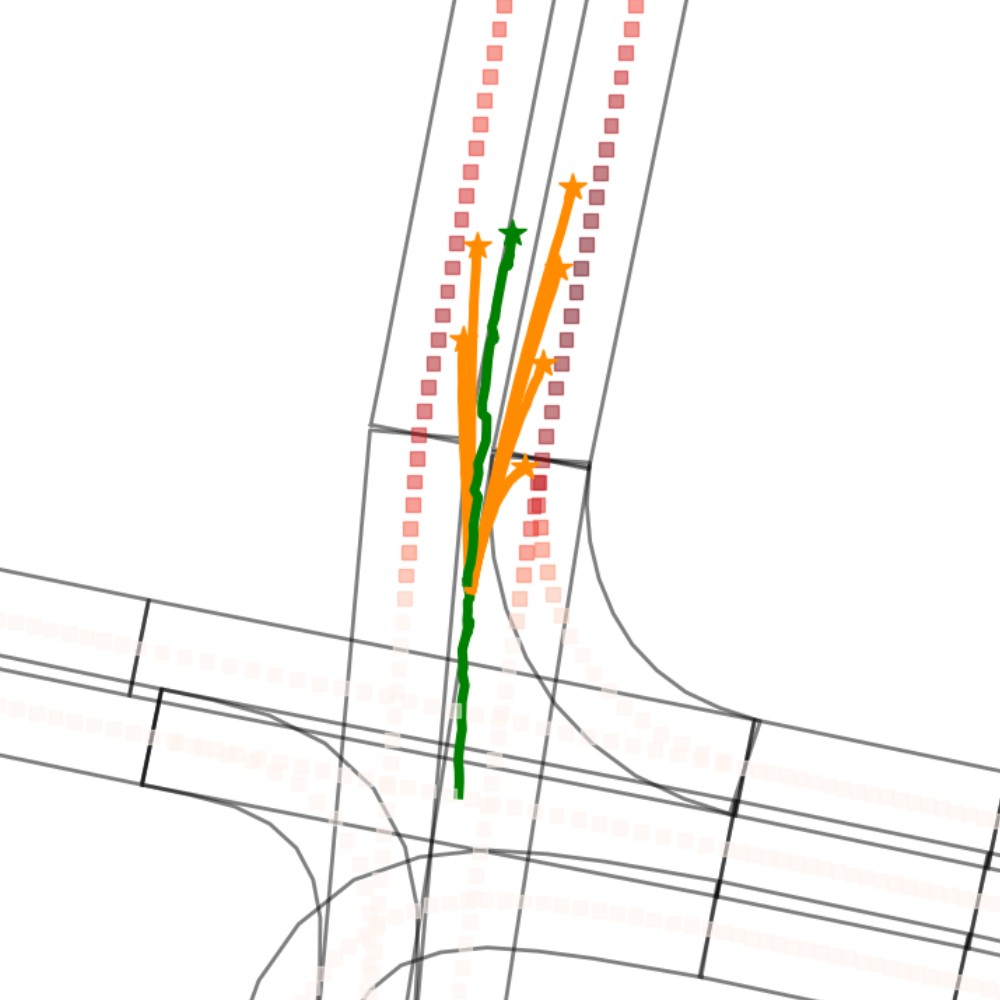}}\vspace{6pt}
\fbox{\includegraphics[width=42mm, height=42mm]{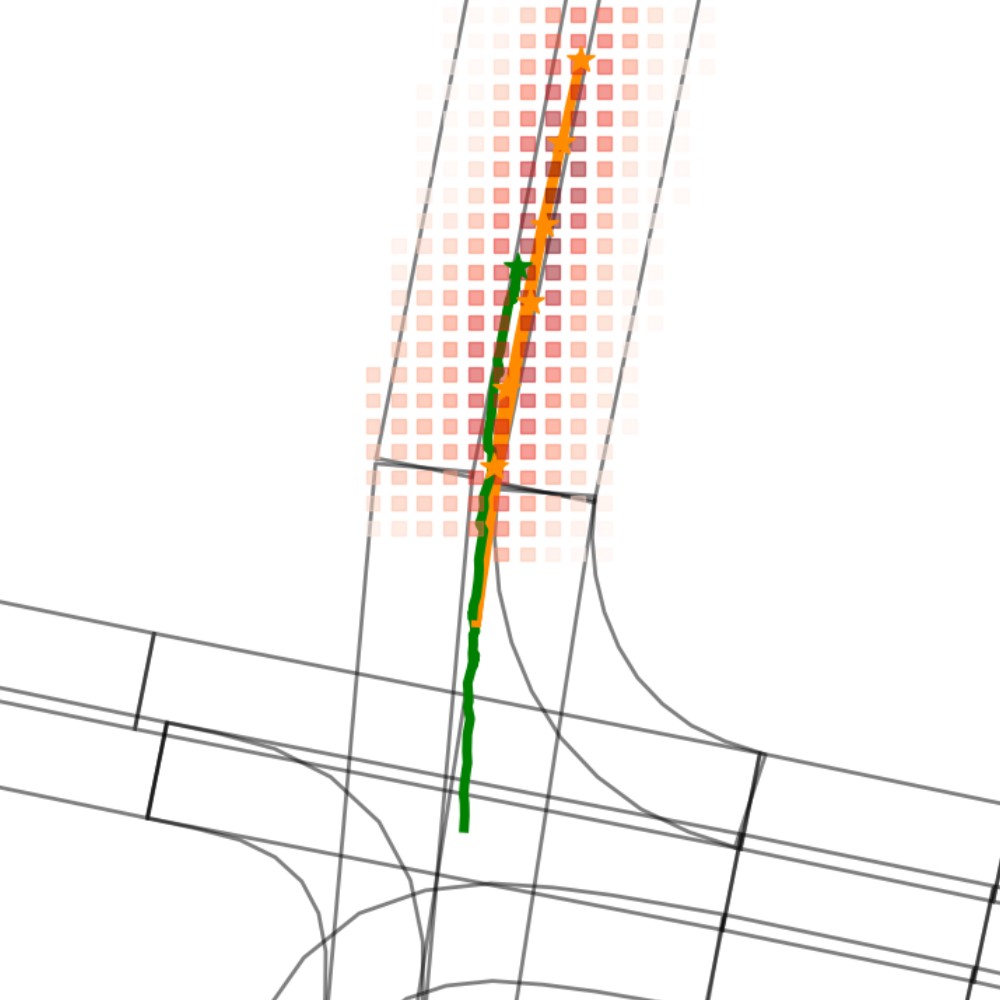}}
\end{minipage}

\caption{\label{appendix:visual}Qualitative comparisons between TNT (the upper row) and our DenseTNT (online, the lower row) on the Argoverse validation set. Here, we choose TNT as a representative for typical goal-based trajectory prediction methods, of which performance heavily depends on the quality of heuristically predefined anchors. Both the anchors of TNT and the heatmaps of DenseTNT are shown in red. Predicted trajectories are shown in orange, and ground truth trajectories are shown in green.}
\end{figure*}

\begin{figure*}[tbh]
\centering
\setlength{\fboxsep}{0pt}

\begin{minipage}[b]{0.23\linewidth}
\fbox{\includegraphics[width=42mm, height=42mm]{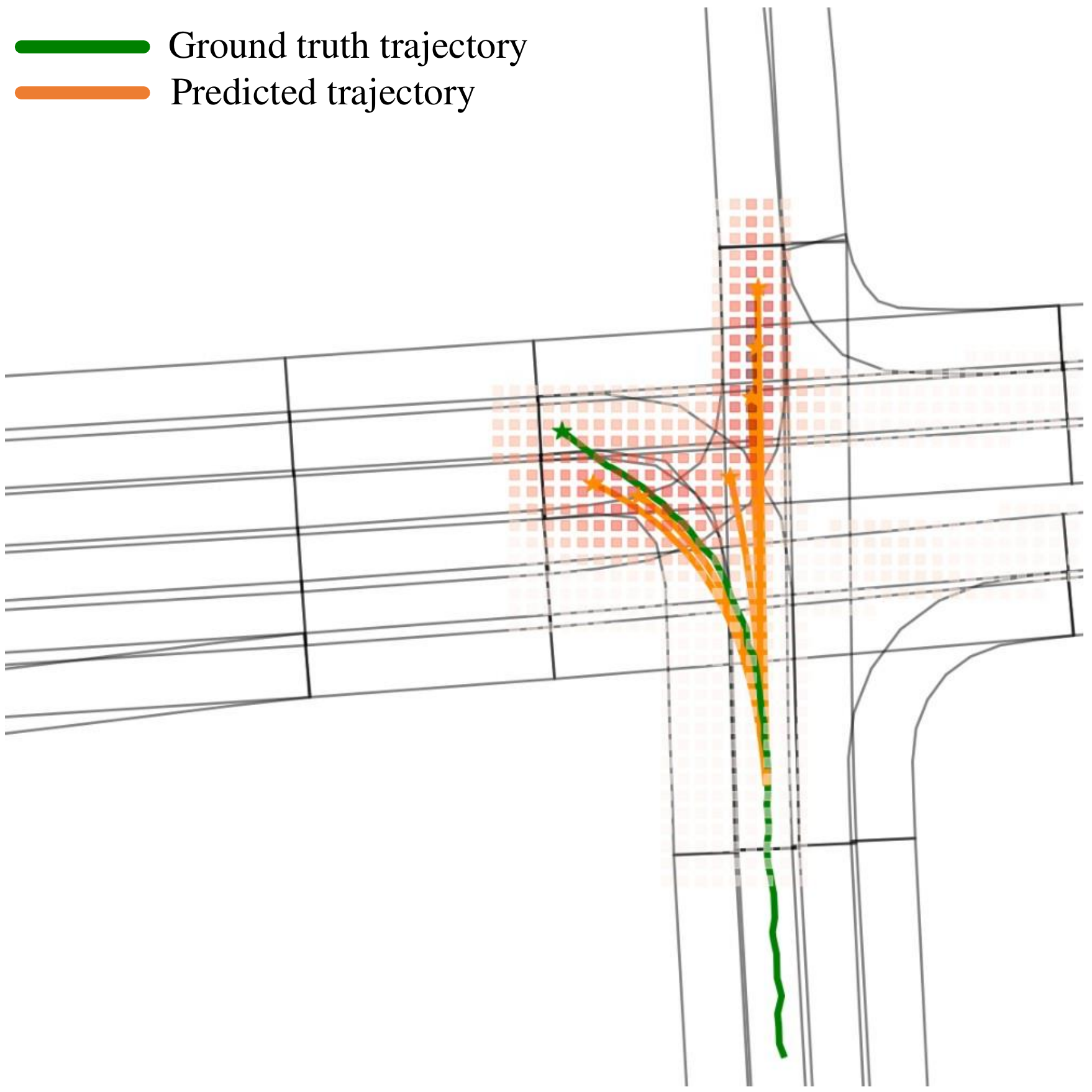}}\vspace{6pt}
\fbox{\includegraphics[width=42mm, height=42mm]{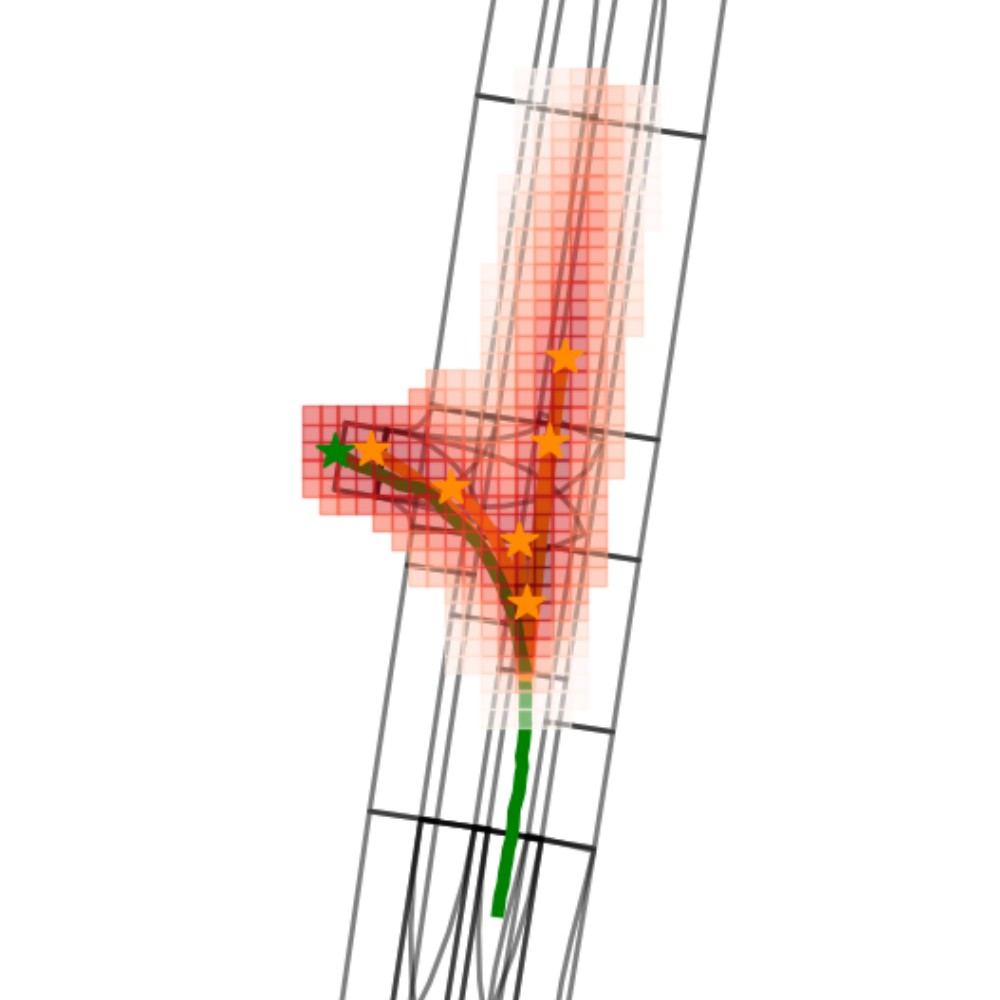}}
\end{minipage}
\hspace{4pt}
\begin{minipage}[b]{0.23\linewidth}
\fbox{\includegraphics[width=42mm, height=42mm]{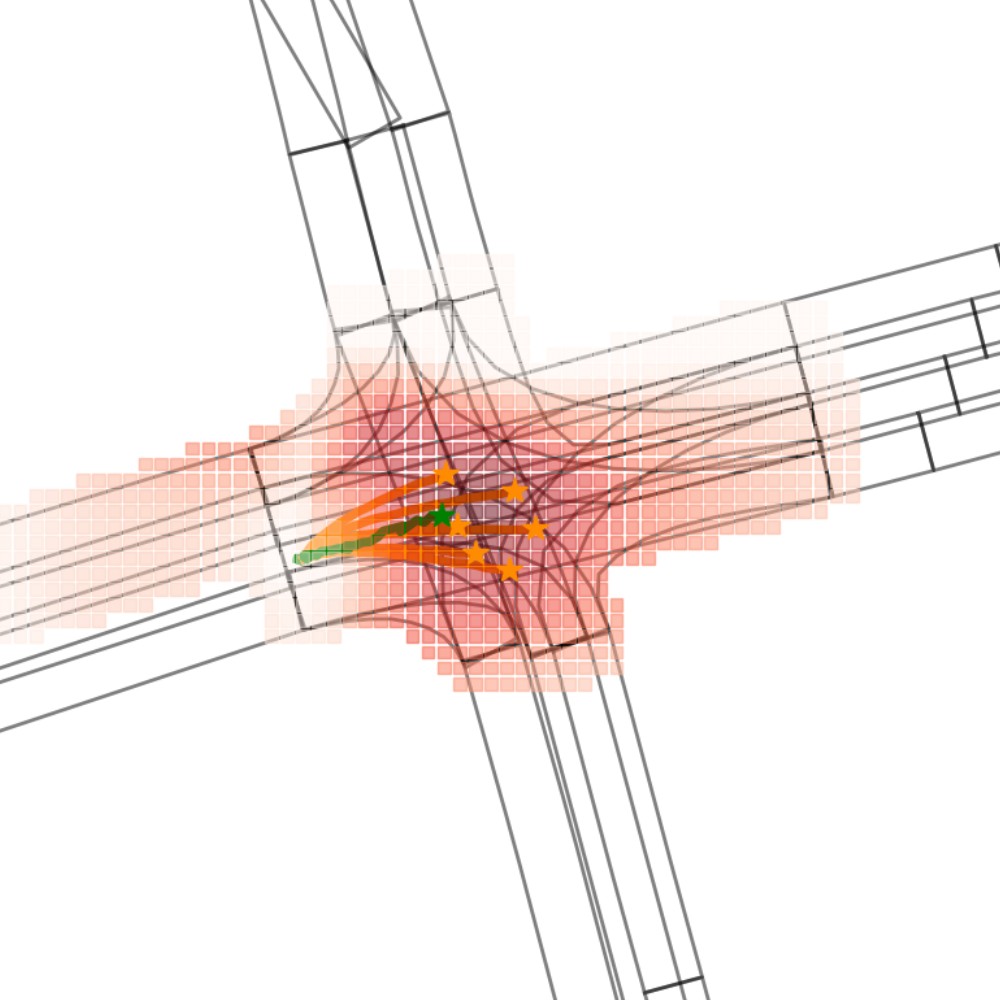}}\vspace{6pt}
\fbox{\includegraphics[width=42mm, height=42mm]{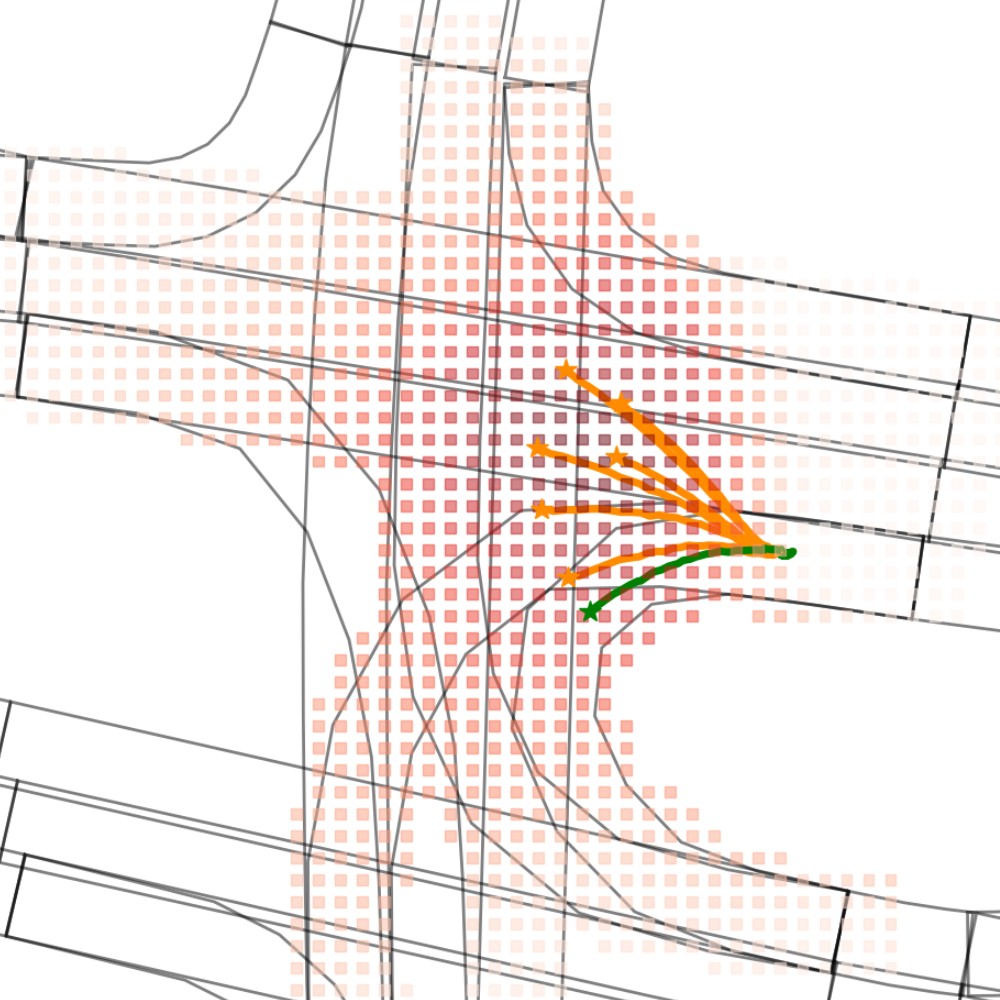}}
\end{minipage}
\hspace{4pt}
\begin{minipage}[b]{0.23\linewidth}
\fbox{\includegraphics[width=42mm, height=42mm]{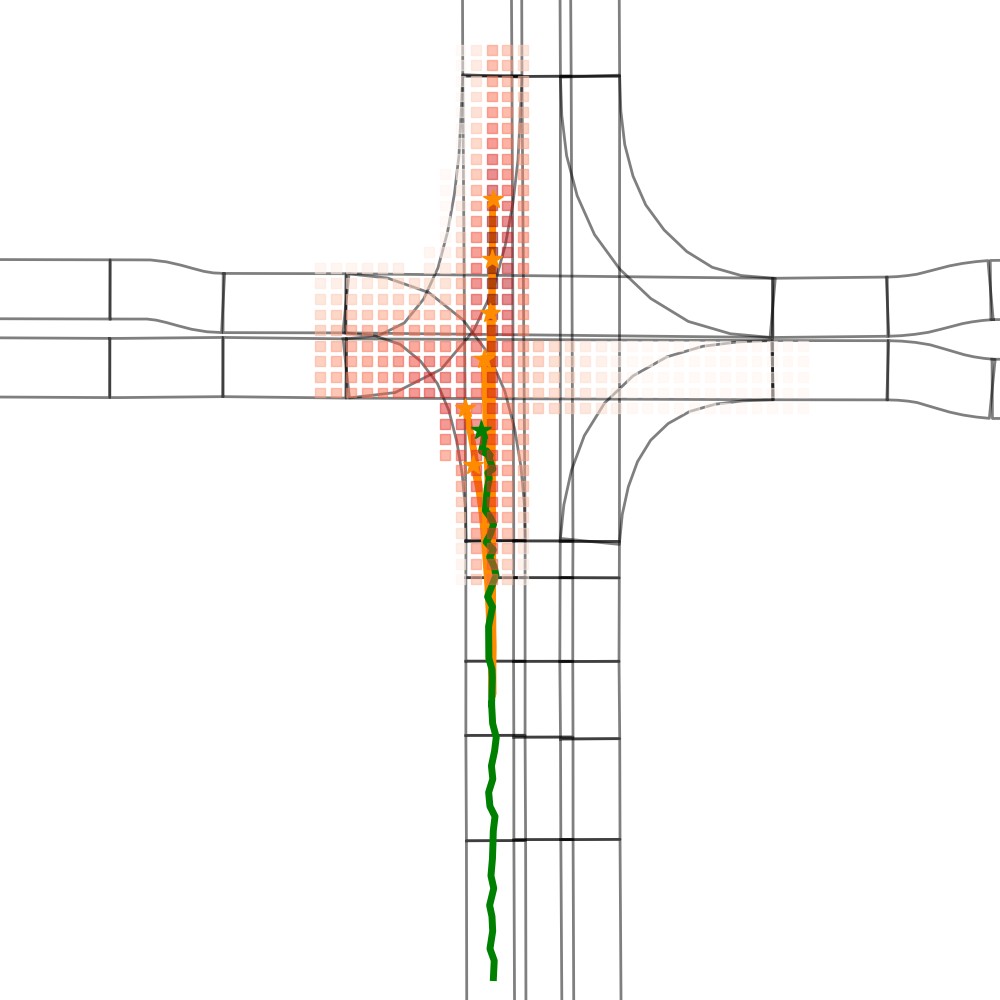}}\vspace{6pt}
\fbox{\includegraphics[width=42mm, height=42mm]{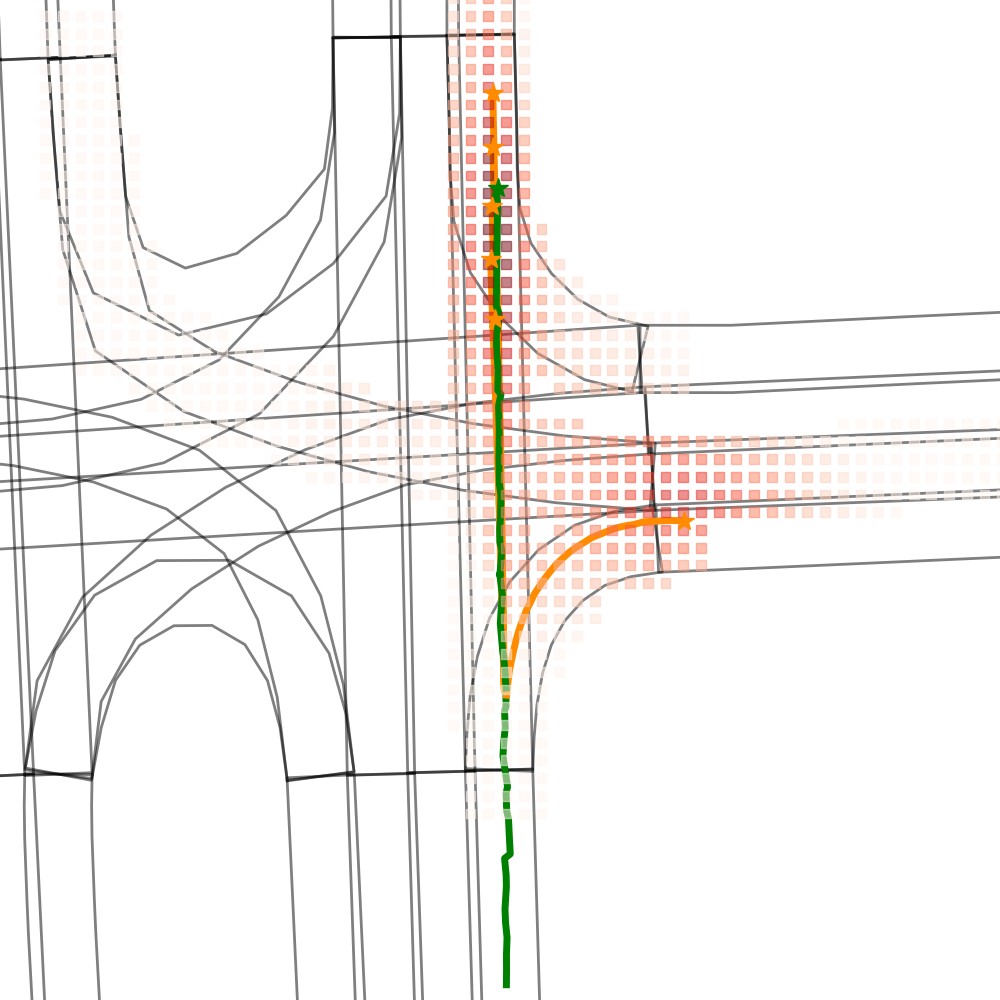}}
\end{minipage}
\hspace{4pt}
\begin{minipage}[b]{0.23\linewidth}
\fbox{\includegraphics[width=42mm, height=42mm]{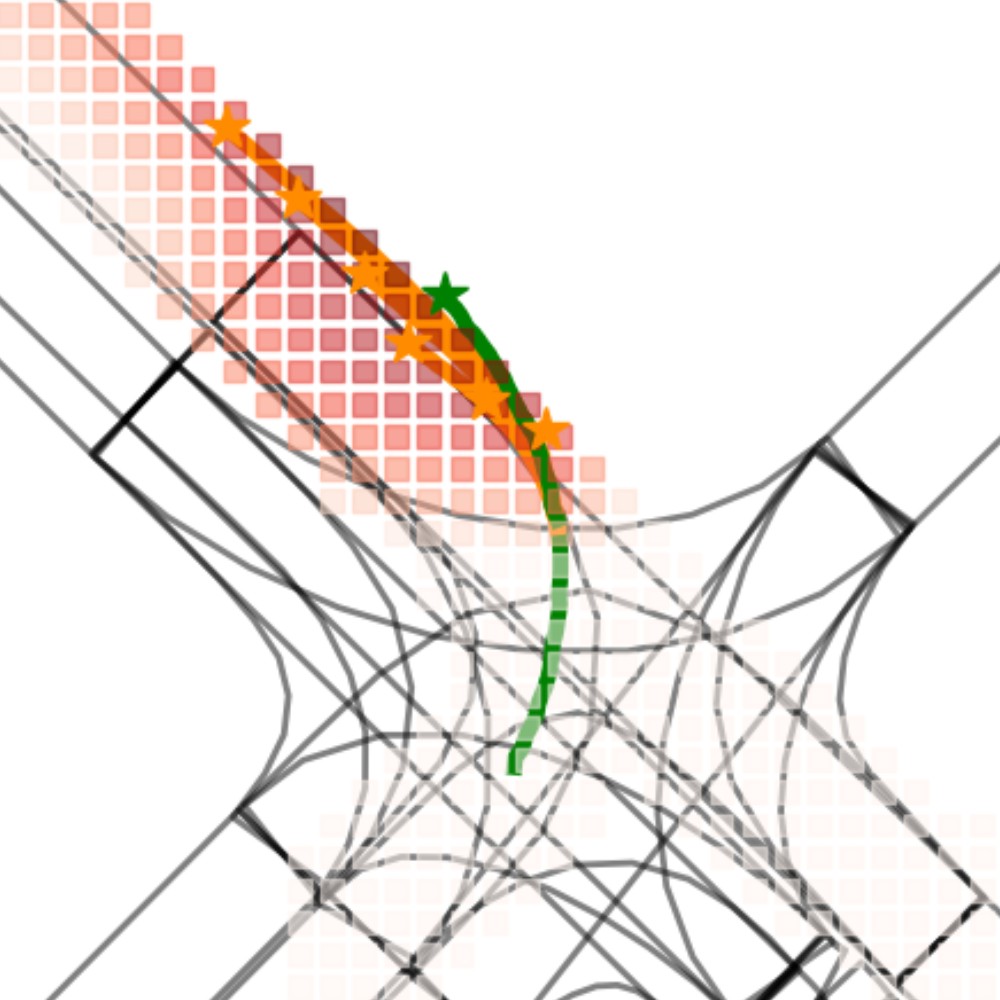}}\vspace{6pt}
\fbox{\includegraphics[width=42mm, height=42mm]{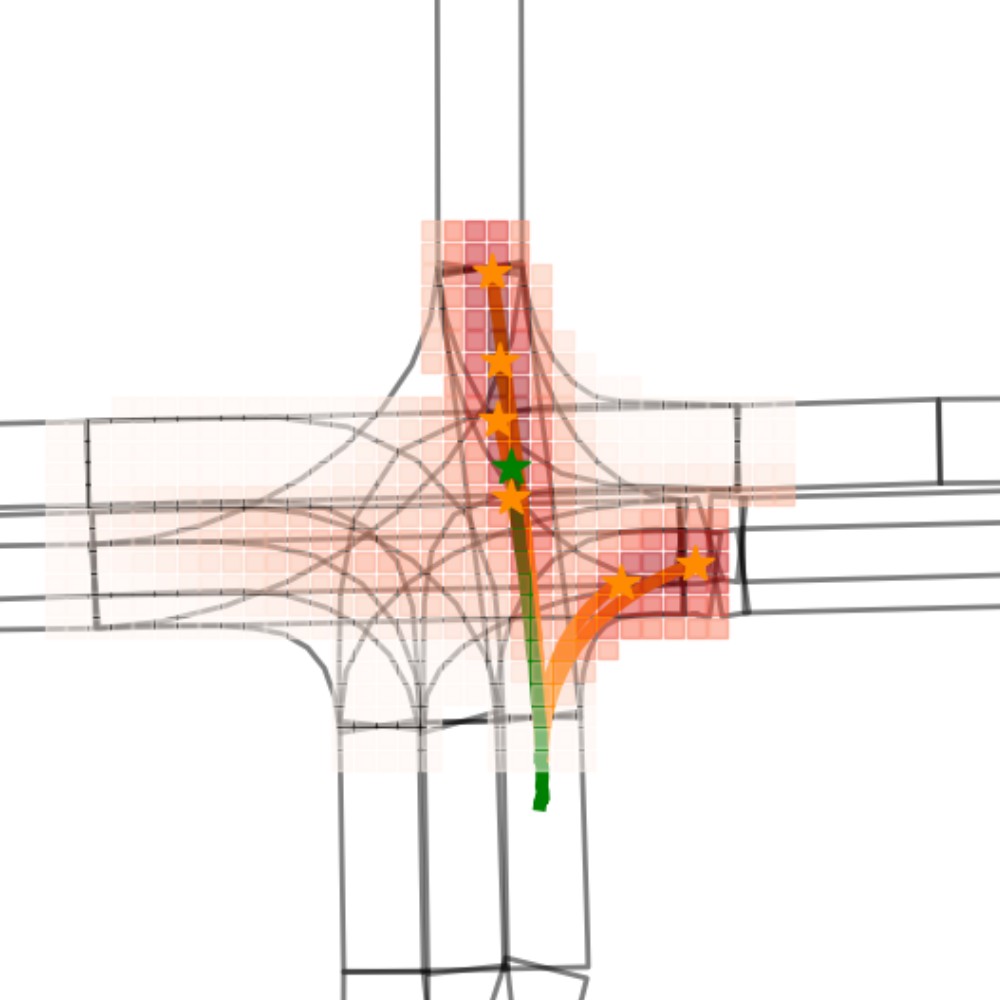}}
\end{minipage}

\caption{\label{more_visual}Qualitative results of DenseTNT (online) in diverse traffic scenarios on the Argoverse validation set. Dense predicted heatmaps are shown in red, predicted goal sets and corresponding trajectories are shown in orange, ground truth trajectories are shown in green.}
\end{figure*}

Figure~\ref{appendix:visual} shows some representative comparisons with typical goal-based trajectory prediction methods, of which performance heavily depends on the quality of heuristically predefined anchors.
We also provide more qualitative results in diverse traffic scenarios on the Argoverse validation set in Figure~\ref{more_visual}. The probability distribution of the final position in some cases is pretty diverse, and it is difficult for NMS to handle well.

\end{document}